\documentclass[journal,twoside,web]{ieeecolor}
\usepackage{generic}
\usepackage{cite}
\usepackage{amsmath,amssymb,amsfonts}
\usepackage{algorithmic}
\usepackage{graphicx}
\usepackage{textcomp}
\usepackage{epsfig}
\usepackage{mathtools}
\usepackage{comment}
\usepackage{soul}
\usepackage[utf8]{inputenc}
\usepackage[flushleft]{threeparttable}
\usepackage{booktabs}
\usepackage{multirow}
\usepackage{color}
\usepackage{float}
\usepackage{colortbl}
\usepackage{courier}

\DeclareMathOperator*{\argmax}{arg\,max}

\makeatletter
\def\fnum@figure{\textcolor{subsectioncolor}{\sf Fig.~\thefigure}}
\def\fnum@table{\textcolor{subsectioncolor}{\sf TABLE~\thetable}}
\makeatother

\def\BibTeX{{\rm B\kern-.05em{\sc i\kern-.025em b}\kern-.08em
    T\kern-.1667em\lower.7ex\hbox{E}\kern-.125emX}}

\definecolor{Gray}{gray}{0.85}

\makeatletter
\def\footnoterule{\kern-3\p@
  \hrule \@width 2in \kern 2.6\p@} 
\makeatother
\newcommand{\copyrightnotice}[1]{{%
  \renewcommand{\thefootnote}{}
  \footnotetext[0]{#1}%
}}

\begin{document}
\bstctlcite{IEEEexample:BSTcontrol}
\title{A Human-Centered Machine-Learning Approach for Muscle-Tendon Junction Tracking in Ultrasound Images}
\author{Christoph Leitner, Robert Jarolim, Bernhard Englmair, Annika Kruse, Karen Andrea Lara Hernandez, Andreas Konrad, Eric Su, Jörg Schröttner, Luke A. Kelly, Glen A. Lichtwark, Markus Tilp and Christian Baumgartner \vspace{-1.5cm}
\thanks{The experimental works and cloud deployments of the present study were supported by Google Cloud infrastructure. Labelbox provided a cloud tool for data annotation.}
\thanks{Christoph Leitner, Bernhard Englmair, Karen Andrea Lara Hernandez, Jörg Schröttner and Christian Baumgartner work at the Institute of Health Care Engineering with European Testing Center for Medical Devices, Graz University of Technology, 8010 Graz, Austria (e-mail: christoph.leitner@tugraz.at).}
\thanks{Robert Jarolim works at the Institute of Physics, University of Graz, 8010 Graz Austria.}
\thanks{Karen Andrea Lara Hernandez works at the Department of Biomedical Engineering, Galileo University, Guatemala City, Guatemala}
\thanks{Christoph Leitner, Annika Kruse, Andreas Konrad and Markus Tilp work at the Institute of Human Movement Science, Sport and Health, University of Graz, 8010 Graz, Austria.}
\thanks{Eric Su, Luke A. Kelly and Glen A. Lichtwark work at the School of Human Movement and Nutrition Sciences, University of Queensland, Brisbane, Australia}
\thanks{Christoph Leitner and Robert Jarolim contributed equally to this work.}}

\maketitle
\copyrightnotice{Copyright (c) 2021 IEEE. Personal use of this material is permitted. Permission from IEEE must be obtained for all other uses, in any current or future media, including reprinting/republishing this material for advertising or promotional purposes, creating new collective works, for resale or redistribution to servers or lists, or reuse of any copyrighted component of this work in other works..}

\begin{abstract}
Biomechanical and clinical gait research observes muscles and tendons in limbs to study their functions and behaviour. Therefore, movements of distinct anatomical landmarks, such as muscle-tendon junctions, are frequently measured. We propose a reliable and time efficient machine-learning approach to track these junctions in ultrasound videos and support clinical biomechanists in gait analysis. In order to facilitate this process, a method based on deep-learning was introduced. We gathered an extensive dataset, covering 3 functional movements, 2 muscles, collected on 123 healthy and 38 impaired subjects with 3 different ultrasound systems, and providing a total of 66864 annotated ultrasound images in our network training. Furthermore, we used data collected across independent laboratories and curated by researchers with varying levels of experience. For the evaluation of our method a diverse test-set was selected that is independently verified by four specialists. We show that our model achieves similar performance scores to the four human specialists in identifying the muscle-tendon junction position. Our method provides time-efficient tracking of muscle-tendon junctions, with prediction times of up to 0.078 seconds per frame (approx. 100 times faster than manual labeling). All our codes, trained models and test-set were made publicly available and our model is provided as a free-to-use online service on https://deepmtj.org/.
\end{abstract}

\begin{IEEEkeywords}
Attention Mechanism, Anatomical Landmark Detection, Convolutional Neural Network, Domain Generalization, Feature Extraction, Label Noise, Locomotion, Myotendinous Junction, Probability Map, Segmentation, Sequential Learning, Soft Labeling, U-Net.
\end{IEEEkeywords}

\section{INTRODUCTION}
\label{sec:introduction}
\begin{figure}[hbt!]
  \centering
  \includegraphics[width=0.35\textwidth, clip, trim={0 0 0 0}]{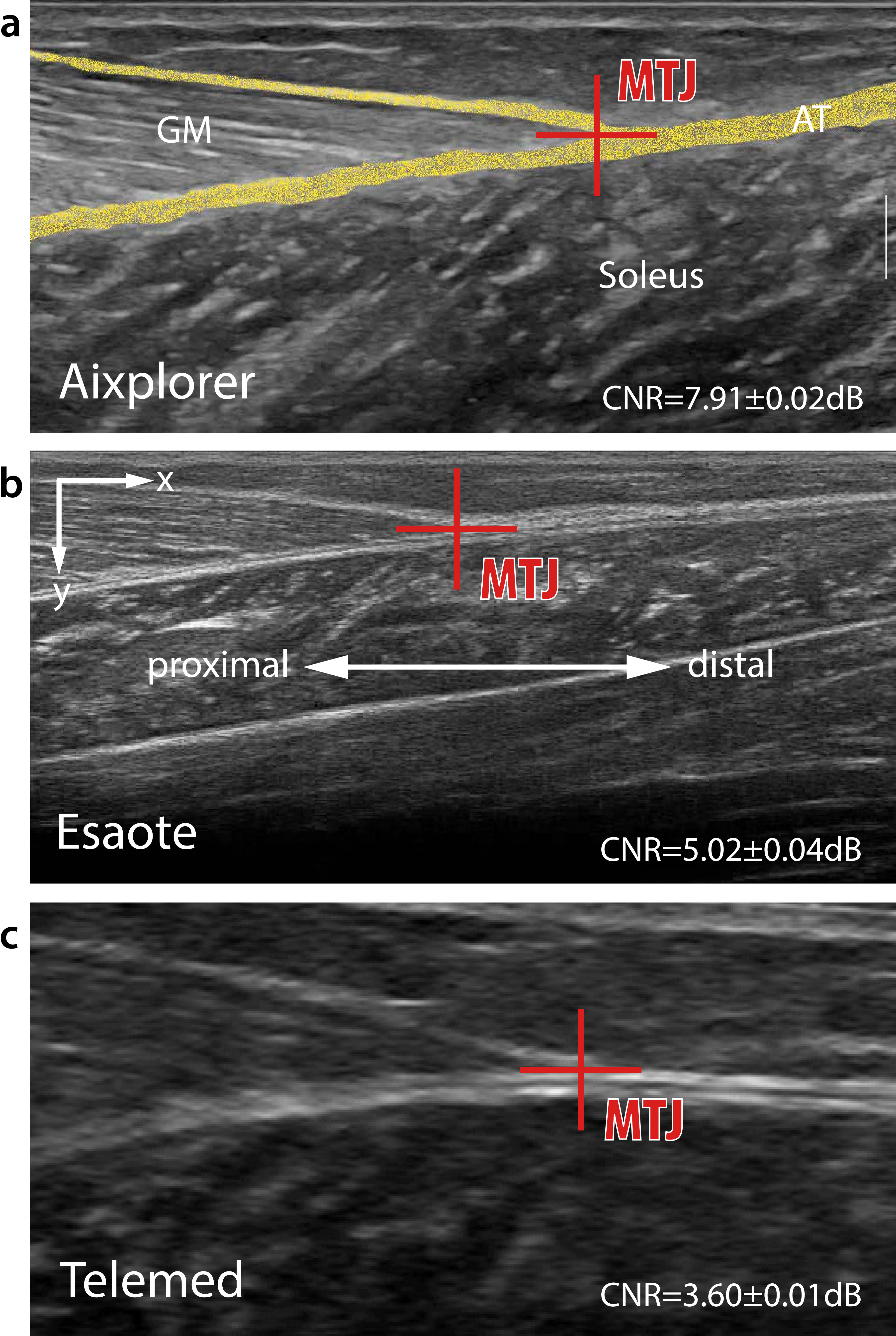}
  \vspace{0.5cm}
  \caption{Three examples of the MTJ in the medial gastrocnemius (MG) muscle-tendon unit, recorded with three different instruments. The MTJ is indicated by a red cross. We specify contrast-to-noise ratios (CNR) for all considered instruments. The video frame in Figure \textbf{a} was collected with an Aixplorer V6 US system (Aixplorer). This figure also shows the embedding (yellow color) of the MTJ in the triceps surae muscle-tendon unit (MG, lateral gastrocnemius (LG, not shown), Soleus and Achilles Tendon (AT)). The white arrow in Figure \textbf{b} indicates the direction of principal movement of the MTJ from distal to proximal in the x,y-coordinate system. This video frame was collected with an Esaote MyLab 60 US system (Esaote). Figure \textbf{c}  shows an image of the MTJ collected with a Telemed ArtUs US system (Telemed).}
  \label{fig:MTJ}
  \vspace{0.5cm}
\end{figure}
%
%
%
%
\begin{table*}[ht!]
    \caption{Specifications of included datasets.}
    \label{tab:studies}
    \setlength{\tabcolsep}{3pt} 
    \begin{center}
    \begin{tabular}{cccccccccccc}
        \toprule
        \multirow{2}{*}{\textbf{Nr.}}
        &\multirow{2}{*}{\textbf{Study}}
        &\textbf{Ultrasound}
        &\multirow{2}{*}{\textbf{Transducer}}
        &\textbf{C. Freq.}
        &\textbf{Frame-rate}
        &\multirow{2}{*}{\textbf{Movement}}
        &\multirow{2}{*}{\textbf{Muscle}}
        &\multirow{2}{*}{\textbf{Subjects (H / I$^\dagger$)}}
        &\textbf{Tot. Nr.}\\
        &
        &\textbf{System}
        &
        &\textbf{[MHz]}
        &\textbf{[Hz]}
        &
        &
        &
        &\textbf{Recordings}\\
        \midrule
        \rule{0pt}{12pt}
        \textbf{1$^{\ddagger}$}
        &\cite{j:Konrad2014:1,j:Konrad2015,j:Konrad2014:2} 
        &Esaote MyLab60 
        &LA923
        &7
        &25
        &MVC, PT  
        &MG 
        &66 (66 / 0)
        &797\\
        \rule{0pt}{12pt}
        \textbf{2$^{\ddagger}$}
        &\cite{j:Kruse2017,j:Kruse2018,j:Kruse2019}
        &Esaote MyLab60
        &LA923
        &7 
        &25
        &MVC, PT
        &MG
        &41 (13 / 28)
        &309\\
        \rule{0pt}{12pt}
        \textbf{3$^\star$}
        &unpub.
        &Telemed ArtUs
        &LV8-5N60-A2
        &8 
        &30-34
        &MVC
        &MG
        &9 (9 / 0)
        &113\\
        \rule{0pt}{12pt}
        \textbf{4$^\|$}
        &unpub.
        &Telemed ArtUs
        &LV8-5N60-A2
        &8 
        &34
        &MVC
        &MG
        &10 (0 / 10)
        &51\\
        \rule{0pt}{12pt}
        \multirow{2}{*}{\textbf{5$^\mathsection$}}
        &\multirow{2}{*}{this study}  
        &Telemed ArtUs
        &LV8-5N60-A2
        &8
        &60-80
        &MVC, PT, RUN
        &\multirow{2}{*}{MG, LG}
        &\multirow{2}{*}{35 (35 / 0)}
        &326\\ 
        &
        &Aixplorer V6
        &SL10-2
        &9
        &25
        &MVC, PT
        &
        &
        &151\\ 
        \bottomrule
        \addlinespace[4pt]
        &
        &
        &
        &
        &
        &
        &\textbf{\small TOTAL:}
        &\textbf{\small 161 (123 / 38)}
        &\textbf{\small 1747}\\
    \end{tabular}
    \end{center}
    \begin{tablenotes}
        \item MVC ... isometric maximum voluntary contraction, PT ... passive torque movement, RUN ... running, MG ... medial gastrocnemius, LG ... lateral gastrocnemius, $\dagger$ Healthy/Impaired, unpub. ... unpublished.
        \item Ethics commissions (approval numbers): University of Graz Ethics Commission (GZ. 39/2/63 ex 2011/12$^{\ddagger}$), The University of Queensland Human Research Ethics Committee (2018000525$^\star$, 2018000856$^\|$), University of Graz Ethics Commission (GZ. 39/56/63 ex 2018/19$^\mathsection$), Medical University of Graz (EK-Nr. 21-362 ex 09/10$^\mathsection$)
    \end{tablenotes}
    \vspace{-0.3cm}
\end{table*}
%
\IEEEPARstart{D}{uring} human locomotion, muscle-tendon complexes of lower limbs are under cyclic concentric and eccentric stress \cite{j:Komi2000}. Within these units, muscles and tendons have different properties \cite{j:Alexander1991}, contribute differently to external loading \cite{j:Lichtwark2006} and adapt differently to stimuli \cite{j:Magnusson2008}. For instance, imbalances in muscle strength or tendon stiffness may impede efficient interplay during locomotion \cite{j:Roberts1997} or lead to injuries \cite{j:Arampatzis2020}. In clinical populations, knowledge of alterations in muscles and tendons due to short or long-term treatments (e.g., physical therapy or surgeries) is crucial for developing efficient therapeutic strategies \cite{j:Hoesl2020}.

To investigate tissue behaviour in lower limbs (e.g., the triceps surae muscle tendon unit) and to distinguish between individual contributions of muscles and tendons, their junctions are usually visualized using ultrasound (US) imaging (Fig. \ref{fig:MTJ}) while their displacements are tracked with various methods \cite{j:Lee2008, j:Zhou2018, j:Cenni2019, c:LeitnerJarolim2020, j:Krupenevich2021}. The triceps surae (Fig. \ref{fig:MTJ}. a) is a major contributor to human locomotion. It consists of three heads: the medial (MG) and lateral (LG) gastrocnemius as well as the soleus (SO) muscle. Each individual head is connected via a muscle-tendon junction (MTJ) to the Achilles tendon (AT). Thus, the MTJ provides a form and force-locked interconnection between contracting muscles and passively acting tendons \cite{j:Charvet2012}. In US images, the MTJ is clearly visible due to the change of acoustic impedance in muscles and tendons. Moreover, due to its definable maximum displacement and primary longitudinal (distal to proximal) travel direction (Fig. \ref{fig:MTJ}. b), the area of this anatomical feature is covered by standard-sized linear US arrays \cite{j:Leitner2019}. Therefore, this method is widely used and it improves general understanding of muscle-tendon properties and their behaviour in healthy \cite{j:Werkhausen2018} and impaired subjects \cite{j:Barber2017, j:Cenni2019}.

However, musculoskeletal US imaging depends on operators \cite{j:Ohrndorf2010}. In particular, image interpretation requires trained specialists. Moreover, investigating displacements of the MTJ in US images typically needs handcrafted labeling. For this reason, several semi-automatic and automatic methods to track MTJs have been proposed \cite{j:Lee2008,j:Zhou2018,j:Cenni2019,j:Kharazi2020}. Image analysis in biomechanical and clinical US studies relies largely on computer vision algorithms \cite{j:Hooren2020}. Applied on noisy, real-world US motion data, these optical-flow or matching based methods are prone to errors \cite{proc:Sukhwan2001}. This is often due to low frame-rate recordings or poor image qualities of standard medical US systems \cite{j:Leitner2019}. One common parameter to quantify US image quality is the contrast-to-noise-ratio (CNR) \cite{j:Ng2011}.

Recently, machine learning solutions for automatic detection and tracking of musculoskeletal features in biomechanical applications have been developed \cite{j:Cronin2020}. These methods improve performance because they can learn to extract salient features, such as anatomical landmarks, directly from annotated input images. Therefore, a neural network is trained to find a mapping between input images and manually set labels \cite{Goodfellow-et-al-2016, lecun2015deep}. With a sufficiently large dataset, neural networks can successfully map novel data (generalization). During training with real world images, the network learns to neglect noise or instrumental errors, yielding robust and accurate results compared with classical computer vision applications \cite{c:Englmair2020}. Leitner \textit{et al.} \cite{c:LeitnerJarolim2020} for example, used data from an Esaote US system (Esaote SpA, Genoa, Italy) and a ResNet model architecture \cite{c:He2016} with an attention mechanism \cite{j:Jetley2018} to investigate MTJ predictions on 107 subjects using 7200 manually annotated labels. They found that an inclusion of healthy and impaired patients into the training dataset improved overall performance of their model. Krupenevich \textit{et al.} \cite{j:Krupenevich2021} focused their work on the trackability of MTJs across several isometric movements and complex functional tasks such as walking. They trained a MobileNetV2 \cite{c:Sandler2018} architecture on 1200 manually annotated ground truth labels, collected from 15 subjects that were walking, with a Telemed US system (Telemed UAB, Vilnius, Lithuania). 

These newly emerging machine-learning applications for MTJ tracking show that deep neural networks provide strong performance in identifying the exact MTJ positions in US images, even for small training datasets, and independent of subjects and movements. However, there is still lack of evidence on how these algorithms perform on noisy inter-laboratory, inter-observer data and may be generalized to diverse settings. Furthermore, previous MTJ tracking neural network models were evaluated on inaccessible test-sets and labels of test-set and training dataset were identified by the same person. This neglects unavoidable positional variations of different observers and introduces potential bias. In particular, machine learning benchmarks need to include more than one clinical specialist to generate reliable reference test-set labels \cite{j:Hernandez2021, j:Li2020, j:Titano2018} with low noise \cite{j:Karimi2020}. Moreover, predictions across multiple-domains (e.g. data collected from different instruments) are key in generalizing machine learning algorithms \cite{j:Zech2018}. For example, deep-learning has shown excellent performance if training and test-set data are drawn from the same underlying distribution. However, large domain shifts in data (e.g. using data from machines of different vendors) often cause significant performance impairments. In case of the proposed MTJ tracker by Leitner \textit{et al.} \cite{c:LeitnerJarolim2020} and Krupenevich \textit{et al.} \cite{j:Krupenevich2021}, evaluation and training dataset come from the same US instrument. Therefore, these networks might fail to provide similar performance on datasets obtained from other US machine types.

In this work, we present a novel deep-learning approach for the detection of MTJs in ultrasound images. We curate a large and diverse training dataset, in order to provide a universal MTJ detection method, independent of the used US instrument, movement, muscle region or noise coming from experimental setups (Sect. \ref{sec:methods.data}). We use a deep neural network with U-Net architecture and attention mechanism to predict the position of the MTJ as a probability density function (Sect. \ref{sec:methods.model}). An objective test-set was created and curated by four independent specialist to evaluate the average deviation of our model from specialist labels (Sect. \ref{sec:results}). In addition, we estimate the generalization to novel datasets and discuss the capabilities of our method (Sect. \ref{sec:generalization}).

%
%
%
%
\begin{table}[t]
    \caption{Distribution of training set labels}
    \label{tab:groundtruth}
    \setlength{\tabcolsep}{3pt} 
    \hspace*{-1.3cm} 
    \begin{center}
    \begin{tabular}{lccc c r}
        \toprule
        &
        \multicolumn{3}{c}{MG (LG)}
        &
        &
        \\
        \cline{2-4}\\
        &\textbf{Aixplorer$^{\dagger}$} 
        &\textbf{Esaote$^{\dagger\dagger}$}
        &\textbf{Telemed$^{\dagger\dagger\dagger}$}
        &
        &\textbf{TOTAL}\\ 
        \midrule
        \rule{0pt}{12pt}MVC
        &2020 (1216)
        &6295 (0)
        &3094 (282)
        &
        &12907\\
        \rule{0pt}{12pt}PT
        &2184 (1868)
        &3826 (0)
        &38274 (2900)
        &
        &49052\\
        \rule{0pt}{12pt}RUN
        &0 (0)
        &0 (0)
        &4905 (0)
        &
        &4905\\
        \bottomrule
        \rule{0pt}{12pt}\textbf{TOTAL} 
        &4204 (3084)
        &10121 (0)
        &46273 (3182)
        &
        &\textbf{66864}\\
    \end{tabular}
    \end{center}
        \begin{tablenotes}
        \item $^{\dagger}$ ... Aixplorer V6, $^{\dagger\dagger}$ ... Esaote MyLab 60, $^{\dagger\dagger\dagger}$ ... Telemed ArtUs, MVC ... isometric maximum voluntary contraction, PT ... passive torque movement, RUN ... running, MG ... medial gastrocnemius, LG ... lateral gastrocnemius.\\
    \end{tablenotes}
    \vspace{-0.5cm}
\end{table}

\section{MATERIALS AND METHODS}
\label{sec:methods}

\subsection{Dataset and Labeling}
\label{sec:methods.data}
We used five different datasets in this study (Table \ref{tab:studies}). Data were collected at the University of Graz, the Graz University of Technology and the University of Queensland between 2014 and 2020 on 123 healthy and 38 impaired individuals. Our research was approved by the responsible ethics committees, and their approval numbers are given in Table \ref{tab:studies}. With 1590 recordings, the isometric maximum voluntary contractions (MVC) and passive torque movements (PT) on the medial gastrocnemius (MG) had the largest share in the dataset. A smaller amount of data was collected on the MG during running (48 recordings). The measurements on the lateral gastrocnemius (LG) consist of 109 recordings. The complete and fully anonymous dataset holds 1747 video recordings with a mean length of 19.84 seconds per video. Sequences were captured at frame-rates of 30 frames per second (fps) for studies with an Aixplorer V6 (SuperSonic Imagine, Aix-en-Provence, France) US system (Aixplorer, Fig. \ref{fig:MTJ}. a), 25 fps for studies with the Esaote MyLab60 system (Esaote, Fig. \ref{fig:MTJ}. b), and 30-80 fps for the Telemed ArtUs US (Telemed, Fig. \ref{fig:MTJ}. c), respectively.

\subsection*{Training Dataset Labels}
The position of the MTJ was identified as the most distal insertion of the muscle into the free tendon (Fig. \ref{fig:MTJ}). Datasets 1, 2, 4 and 5 (Table \ref{tab:studies}) were annotated by senior investigators ($>$3 years of research experience, conducted $>$2 subject trials investigating the MTJ in the past 2 years). Three cycles of reviews on the labels were conducted by the same annotator, with a minimum interval of 2 days between reviews. Dataset 3 (Table \ref{tab:studies}) was annotated by a junior investigator who was carefully trained to annotate US data of the MTJ. Labels were reviewed by a senior investigator ($>$10 years of research experience, conducted $>$4 subject trials investigating the MTJ in the past 2 years).

\begin{figure*}[t!]
  \centering
  \vspace{-0.5cm}
  \includegraphics[width=0.9\textwidth, clip, trim={0 0 0 0}]{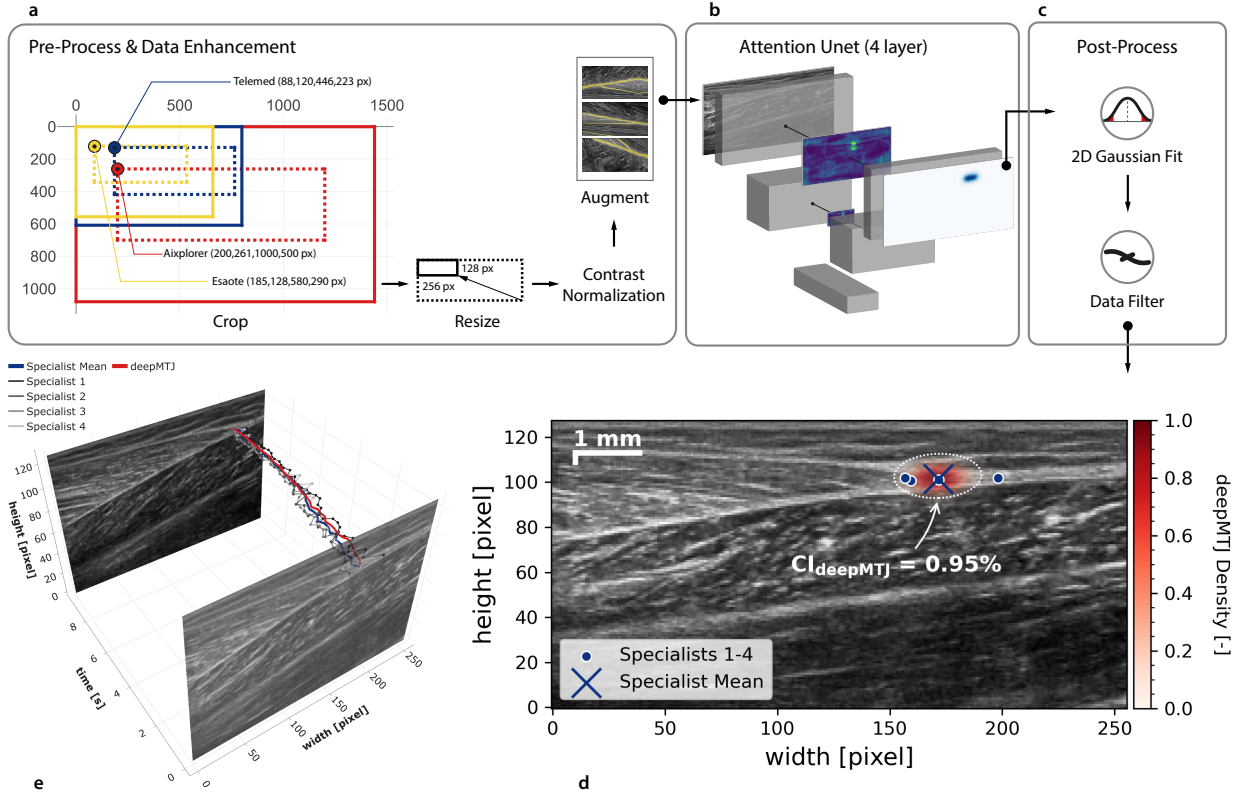}
  \vspace{-0.1cm}
  \caption{This figure demonstrates the processing workflow of our machine-learning approach. In \textbf{a} we show pre-processing steps applied to data such as cropping, resizing and rescaling. Furthermore, we have included image augmentation to account for noise occurring in the experimental data collection. \textbf{b} shows a U-Net architecture with 4-layers and attention mechanism. Our model processes probability maps with soft labels. Therefore, small positional uncertainties about the MTJ did not cause large losses. Figure \textbf{c} demonstrates that we applied a 2D Gauss fit on predicted maps and evaluated filtered noise coming from our algorithm or large inter-specialist variability. For our data analyses depicted in Figure \textbf{d}, we compared soft labels set by our model (red overlay) with individual specialist labels (blue dots with white edges). The 95\% confidence interval for our model label pixel position is indicated with the white dotted ellipse. We defined the specialist mean position (blue cross) as the target reference label for our performance evaluation. Figure \textbf{e} shows an example of a complete time-series reconstruction of one PT movement included in the test-set. We have indicated specialists (gray color traces), reference labels (blue trace), and the prediction trace of our model (red color).}
  \label{fig:data_network}
\end{figure*}

Table \ref{tab:groundtruth} shows a detailed distribution of all labels over included instruments, muscles and movements. A labeling frequency of 10 frames per video was chosen for dataset 1. Every $5^{th}$ frame was annotated in dataset 2-5. These sampling frequencies prevent strong temporal correlations. In total, our training dataset holds 66864 ground truth labels, covering 3 different movements, collected on two muscles from 123 healthy and 38 impaired subjects, and were recorded on three US systems from different vendors. 

\subsection*{Test-set Labels}
For model testing, we first randomly selected 12 participants from our dataset and then completely separated all related 107 recordings into an independent test data pool. From there we randomly excluded 13 MVCs and 14 PT movements and cut each video file to a length of 10 seconds covering the movement. Consequently, we excluded every $5^{th}$ frame of each individual time-series. This resulted in a number of 1360 frames in the test-set. We refrained from including the running movement in the test-set because diversity (collected on 3 subjects) and the overall share (7.3\%) was low. In total, our test-set holds 1360 ground truth labels, covering 2 different movements, collected on 2 muscles from 12 individual subjects, and were recorded on 3 US systems from different vendors. 

The test-set was then annotated by $N = 4$ specialists $S = \{1, 2, 3, 4\}$. Specialists had 2-10 years of experience in biomechanical and clinical research investigating muscles and tendons in 2-9 US studies in the past 2 years. 
The data was labeled using a web-based annotation tool (Labelbox Inc., San Francisco, CA, USA). The manual annotation of the test-set took on average $2.76 \pm1.11$ hours. 

To evaluate specialist performances, we used leave-one-out cross validation. We computed Euclidean distances $\boldsymbol{d}^{k}$ between label positions $\boldsymbol{P}^{k}_{x,y}$ of each individual specialist $k \in S$, and mean label positions $\bar{\boldsymbol{P}}^{j}_{x,y}$ of the remaining three specialists $j \in S$. 
Mean deviations of specialists $\bar{\boldsymbol{d}}^{S}$ and their standard deviations $\boldsymbol{\sigma}^{S}$ were computed over all $k$-folds, and used for model evaluations (Sect. \ref{sec:data_analysis}):
\begin{equation}
\label{eq:sp_mean}
    \begin{gathered}
        \bar{\boldsymbol{d}}^{S} = \frac{1}{N} \sum_{j\neq k} \sqrt{(\boldsymbol{P}^{k}_{x} - \bar{\boldsymbol{P}}^{j}_{x})^2+(\boldsymbol{P}^{k}_{y} - \bar{\boldsymbol{P}}^{j}_{y})^2}
    \end{gathered}
\end{equation}
Absolute standard deviation of specialists $\bar{\boldsymbol{\sigma}}^{S}$ is used as a supporting distance metric for our data analysis (Sect. \ref{sec:data_analysis}) and filtering (Sect. \ref{sec:errorcases}). 

From the four specialist labels $\boldsymbol{P}^{k}_{x,y}$, with $k \in S$, we computed average label positions $\bar{\boldsymbol{P}}^{S}_{x,y}$, in order to obtain the most likely positions of the MTJ. These average specialist positions were used as a reference for model evaluations (Sect. \ref{sec:data_analysis}) and were referred to as reference labels. 

To account for label noise among specialists, the interclass correlation coefficient (ICC) for inter-rater analysis was used. Our calculated ICC is based on a mean-rating ($k^{*} = 3$) and absolute-agreement in a 2-way mixed-effects model. Furthermore, we calculated standard error of mean (SEM) and root-mean-square errors (RMSE) over all $k$-folds.

\subsection{Model}
\label{sec:methods.model}

\subsection*{Data Pre-processing}
\label{sec:methods.model.datapre}
Fig. \ref{fig:data_network}a (left) shows image sizes for each instrument output (Aixplorer, Esaote, Telemed). All images were cropped by an aspect ratio of 2 : 1. Crop rectangles are indicated by dashed lines and crop rectangle sizes (x-coordinate, y-coordinate, width, height) by pixels for each instrument. Origins were set at upper left corners. All frames were resized to 256 by 128 pixels by a third order transformation (Fig. \ref{fig:data_network}a center). 

For our neural network training, we applied additional image augmentation, to counterbalance differences and errors in the experimental setups. During experimental trials using bulky standard medical US probes, or in special environments (e.g. data collection on impaired subjects) transducers are often displaced \cite{j:Leitner2019}. This causes image distortions and can also lead to out-of-plane movements of transducers. Furthermore, images of the same tissue can be mirrored in consecutive trials of the same participant due to changed probe orientations after probe reattachment. For our training, we artificially created similar image distortions by random rotations in ranges of $\pm 20$ degrees, zooming images by scale factors of 0.7 to 1.3, with shearing to maximum angles of 0.2, randomly shifting in x- and y-directions by up to 10\% of image sizes and randomly flipping images vertically and horizontally (Appx. \ref{sec:Appendix}, Fig.\ref{fig:Augmentation}). All images were normalized for zero mean and unit standard deviation before used as input for the neural network (contrast normalization as described in Goodfellow \textit{et al.} \cite{Goodfellow-et-al-2016}). From this pre-processing, we assume that our model is more robust to dissimilarities because of experimental setups and movement artifacts (see Sect. \ref{sec:results}).

Manually set labels in the training dataset denote exact pixel positions of estimated MTJ positions in the image. For our training, we used soft labeling, where we assigned probability values to each image pixel (cf., Mathis \textit{et al.} \cite{j:Mathis2018}). We used probability maps at the same resolution as original images, where we modeled positions of the MTJ by a 2D normal distribution with a covariance of 100 pixels at positions of specialist labels. Therefore, model training is more tolerant in terms of small positional uncertainties, in agreement with uncertainties of manual labels. This also accounts for US images where the MTJ is not visible due to out-of-plane movements of transducers, by assigning no MTJ position in the image. For each probability map, we applied the same random image augmentation and resizing to 256x128 pixels as for input images.

\subsection*{Model Architecture}
A convolutional neural network (CNN) was used to predict a probability map of the MTJ position for a given input image. We built upon a U-Net type architecture \cite{j:Ronneberger2015} that encodes the input image into a feature representation using consecutive convolutional and max-pooling layers. These layers reduced spatial dimensions while increasing network depth. This allowed the network to extract salient features while taking global contexts of images into account (e.g., the characteristic Y shape of the muscle-tendon structure). The feature representation was decoded into the probability map by the same number of consecutive upsampling and convolutional layers as for decoding. Between encoding and decoding of the features, we applied skip connections to correlate spatial information with the extracted image features \cite{j:Ronneberger2015}. We assume that only a fraction of extracted features in the encoder is relevant to determine exact MTJ positions. Consequently, we followed the approach by Oktay et al. \cite{j:Oktay2018} and added an attention mechanism to the skip connections of our model. Here the encoded features were used to create attention maps that enabled the network to focus on more important regions (see Fig. \ref{fig:data_network}b). We used a network depth of 4, with a starting filter number of 64 that we increased by factor 2 after each max pooling layer and decreased by factor 2 after each upsampling layer. The final probability map was obtained from the final convolutional layer with 1 channel and a sigmoid activation function, such that each pixel in the input image was labeled with values within range $[0, 1]$, where 1 denotes high probability for the position of the MTJ at the corresponding pixel position.

\subsection*{Training}
\label{sec:methods.train}
For our model training, we used the full training dataset of 66864 manually annotated images. We used binary cross-entropy as a loss function, where we weighted the 0 class with 0.1, in order to counteract class imbalance. The Adam optimizer was used for parameter optimization with a learning rate of $0.0001$, and decay rates $\beta_1 = 0.9$ and $\beta_2 = 0.999$.

To account for domain generalization on different instrumental data, a sequential learning strategy was used. Within three independent training steps, we sequentially added data from the individual instruments. Here we followed the sequence: 1. \{Telemed\}, 2. \{Telemed, Esaote\} and 3. \{Telemed, Esaote, Aixplorer\}, in correspondence with the size of the individual datasets (largest to smallest). We trained our model for 100 epochs per step and kept the pre-trained model weights from previous steps. After each step, model weights were saved, in order to assess the influence of additional data diversity on the model performance (Sect. \ref{sec:results}). For all other evaluations, we used the final model after the third training step.

\subsection*{Post-processing}
\label{sec:methods.post}

From the obtained probability maps, we estimated MTJ positions by fitting a 2D normal distribution to the output of the neural network. For optimization, 2D Gaussian kernels were fitted for each image by employing a trust region reflective algorithm \cite{j:Branch1999}. To set initial conditions for optimization, we first identified points with the highest probability in US images $\hat{\boldsymbol{P}}^{M}_{x,y} = \argmax(\boldsymbol{p}^{M})$ and then set starting conditions for the Gaussian fit as follows: $\boldsymbol{A} = \max(\boldsymbol{p}^{M})-2*SD$ (amplitude), $(x_0,y_0) = \hat{\boldsymbol{P}}^{M}_{x,y}$ (origin), $\sigma_x = \sigma_y = 5 * \sigma$ (standard deviation in x and y direction), $\theta = d = 0$ (angle and offset). Pixel positions with highest kernel density were identified as MTJ positions $\boldsymbol{P}^{M}_{x,y}$ of our network (Fig. \ref{fig:data_network}d). We found that although the training was performed with a pixel-wise loss, the neural network provided similar probability maps to the ground-truth labels. In most cases, this allows for an unambiguous identification of the MTJ (error-cases are discusses in Sect. \ref{sec:errorcases}).

\subsection{Label Noise Filter}
\label{sec:errorcases}
To account for label noise introduced by our algorithm, as well as by inter-observer variability, we identified three error-case scenarios. Images which are classified in either one of these categories were excluded from the test-set. The first two error-cases are related to predictions of our network. We excluded predictions where labels were either located at image borders which is the case if 

\begin{equation}
\label{eq:errorcase1}
    \begin{gathered}
       \boldsymbol{P}^{M}_{x} \in \left\{\min{\big(\boldsymbol{P}^{M}_{x}}\big), \ \max{(\boldsymbol{P}^{M}_{x}\big)}\right\} 
       \\ or \\
      \boldsymbol{P}^{M}_{y} \in \left\{\min{\big(\boldsymbol{P}^{M}_{y}}\big), \ \max{(\boldsymbol{P}^{M}_{y}\big)}\right\}
    \end{gathered}
\end{equation}
or where the labels are within a 10 pixel padding around borders, and probability scores are low ($\max\big(\boldsymbol{p}^{M}\big)<25\%$). Such cases are given by:

\begin{equation}
\label{eq:errorcase2}
    \begin{gathered}
    \boldsymbol{P}^{M}_{x} \notin \Big[
        \min{\big(\boldsymbol{P}^{M}_{x}}\big) + 10px, \: \max{\big(\boldsymbol{P}^{M}_{x}\big) - 10px}\ \Big]\ 
       \\ or \\
        \boldsymbol{P}^{M}_{y} \notin \Big[\min{\big(\boldsymbol{P}^{M}_{y}}\big) + 10px, \: \max{\big(\boldsymbol{P}^{M}_{y}\big) - 10px}\
      \Big]
      \\ and \\
      \max\big(\boldsymbol{p}^{M}\big)<25\% .
    \end{gathered}
\end{equation}
\vspace{0.8cm}

For the first two cases, we excluded 11 images in total (Appx. \ref{sec:Appendix}, Fig. \ref{fig:EC12}). For these cases, our neural network was not able to identify the position of the MTJ position within the image. This erroneous behaviour occurred in 0.8 \% of cases.

The third error-case relates to inconsistent labels among specialists. 5 images of the test-set (Appx. \ref{sec:Appendix}, Fig.\ref{fig:EC3}) were excluded because MTJ labels of at least three specialist showed distances of $20 \cdot \bar{\sigma}^{S}$ from reference labels $\bar{\boldsymbol{P}}^{S}_{x,y}$. This reduced the test-set by a total of 16 images (1.17 \%). Thus, 1344 video frames were used for further performance analyses.

\vspace{0.1cm}
\subsection{Data Analyses}
\label{sec:data_analysis}
We calculated Euclidean distances $\boldsymbol{d}^{M}$ between the reference label positions $\bar{\boldsymbol{P}}^{S}_{x,y}$ obtained from specialists (Sect. \ref{sec:methods.data}) and label positions from our model $\boldsymbol{P}^{M}_{x,y}$, and compared them to mean deviations of specialists $\bar{\boldsymbol{d}}^{S}$. To evaluate labeling performances between our network and specialists RMSEs, SEMs, as well as mean absolute errors were computed. We assessed the agreement between our model and the reference labels in x- and y-coordinates (Fig. \ref{fig:MTJ} b) using Bland-Altman plots. These comparisons are illustrated by differences between pairs of measurements (${d}^{M}_{x}$, ${d}^{M}_{y}$) as a function of the normalized mean measures (normalization to instrument image sizes). Furthermore, we introduced a tolerance distance given as a multiple $n^{*}$ of mean standard deviations $\bar{\sigma}^{S}$ from the reference labels $\bar{\boldsymbol{P}}^{S}_{x,y}$. We estimate the number of correctly identified samples by individual specialists and our model with increasing distance form the reference labels.

\vspace{0.1cm}
\subsection*{Comparison with other MTJ trackers}
We compared our model to recent MTJ tracking algorithms by computing RMSE for each method based on our test-set. We considered the semi-supervised computer vision algorithm from Cenni \textit{et al.} \cite{j:Cenni2019} and deep-learning approaches from Leitner \textit{et al.} \cite{c:LeitnerJarolim2020} and Krupenevich \textit{et al.} \cite{j:Krupenevich2021}. For the algorithm of Cenni \textit{et al.} MTJ positions in the first frame of each video are needed to start predictions. Therefore, we used the reference labels $\bar{\boldsymbol{P}}^{S}_{x_{1,i},y_{1,i}}$ of the first video frame, from the $i^{th}$ video file in the test-set, to mark starting positions for optical flow calculations. MTJ trackers from Leitner \textit{et al.} \cite{c:LeitnerJarolim2020} and Krupenevich \textit{et al.} \cite{j:Krupenevich2021} are based on CNN architectures. We used their provided trained models for predictions on our test-set. For these tracking solutions, evaluation and training datasets come from the same US instruments. To evaluate differences in image qualities of US machine types, and to estimate possible domain specificity of models, CNRs (Fig. \ref{fig:MTJ}) were calculated with the method described in \cite{c:Leitner2021}.

\vspace{0.1cm}
\subsection{Open Source Cloud Deployment}
We implemented a cloud deployment (Google LLC, Mountain View, CA, USA) of our model and provide public access to our software-as-a-service via a web application. To comply with strict data security standards, each platform user receives a private and personal 128-bit Universally Unique Identifier (UUID) to access and interact with the uploaded data. Video files are temporarily stored on cloud infrastructure for prediction. User data is not stored on servers beyond calculations and deleted after job completion or in case of lost connections. Our web application is entirely open-source and publicly available at https://deepmtj.org.

%
%
%
%
\setlength{\tabcolsep}{0.32em}
\begin{table}[t!]
\caption{Quantitative summary of results}
\label{tab:Results}
\begin{center}
    \vspace{-0.2cm} 
    \renewcommand{\arraystretch}{1.20}
    \begin{tabular}{ccccc}
        \toprule
           \multirow{2}{*}{\textbf{Annotator}}
            &\multirow{2}{*}{\textbf{Type}}
            &\textbf{RMSE (RMSE\textsubscript{DS})}    
            &\textbf{SEM}
            &\textbf{time\textsuperscript{*}}\\
            &
            &[mm]
            &[mm]
            &[min]\\
        \midrule
            \cellcolor{Gray}Specialist\textsuperscript{$\ddagger$} 
            &\cellcolor{Gray}Human
            &\cellcolor{Gray}$5.02 \pm0.62$   
            &\cellcolor{Gray}$0.11 \pm0.02$
            &\cellcolor{Gray}$165.6 \pm66.6$\\
            Cenni et al. \cite{j:Cenni2019} 
            &CV
            &10.62
            &0.18
            &8.01\\
            Leitner et al. \cite{c:LeitnerJarolim2020}    
            &ML
            &7.92 (5.57\textsuperscript{$\dagger$})
            &0.14
            &2.58\\
            Krupenevich et al. \cite{j:Krupenevich2021}    
            &ML
            &26.55 (12.06\textsuperscript{$\dagger\dagger$})
            &0.40
            &3.78\\
            \textbf{This work:} 
            &\textbf{ML}
            &\textbf{4.89}
            &\textbf{0.10}
            &\textbf{1.79}\\
        \bottomrule
    \end{tabular}
    \begin{tablenotes}\footnotesize
    \item RMSE ... root mean square error, RMSE\textsubscript{DS} ... domain specific root mean square error, SEM ... standard error of mean, CV ... computer vision, ML ... machine learning
    \item \textsuperscript{$\ddagger$}values are given as mean and standard deviation over all $N=4$ specialists.
    \item \textsuperscript{$\dagger$} only Esaote test-set data considered.
    \item \textsuperscript{$\dagger\dagger$} only Telemed test-set data considered.
    \item \textsuperscript{*}models were tested on a NVIDIA GeForce RTX 2070 graphical processing unit.
    \end{tablenotes}
\end{center}
\vspace{-0.3cm}
\end{table}
\vspace{0.5cm}

\begin{figure}[h!]
  \centering
  \includegraphics[width=0.45\textwidth, clip, trim={0 0 0 0}]{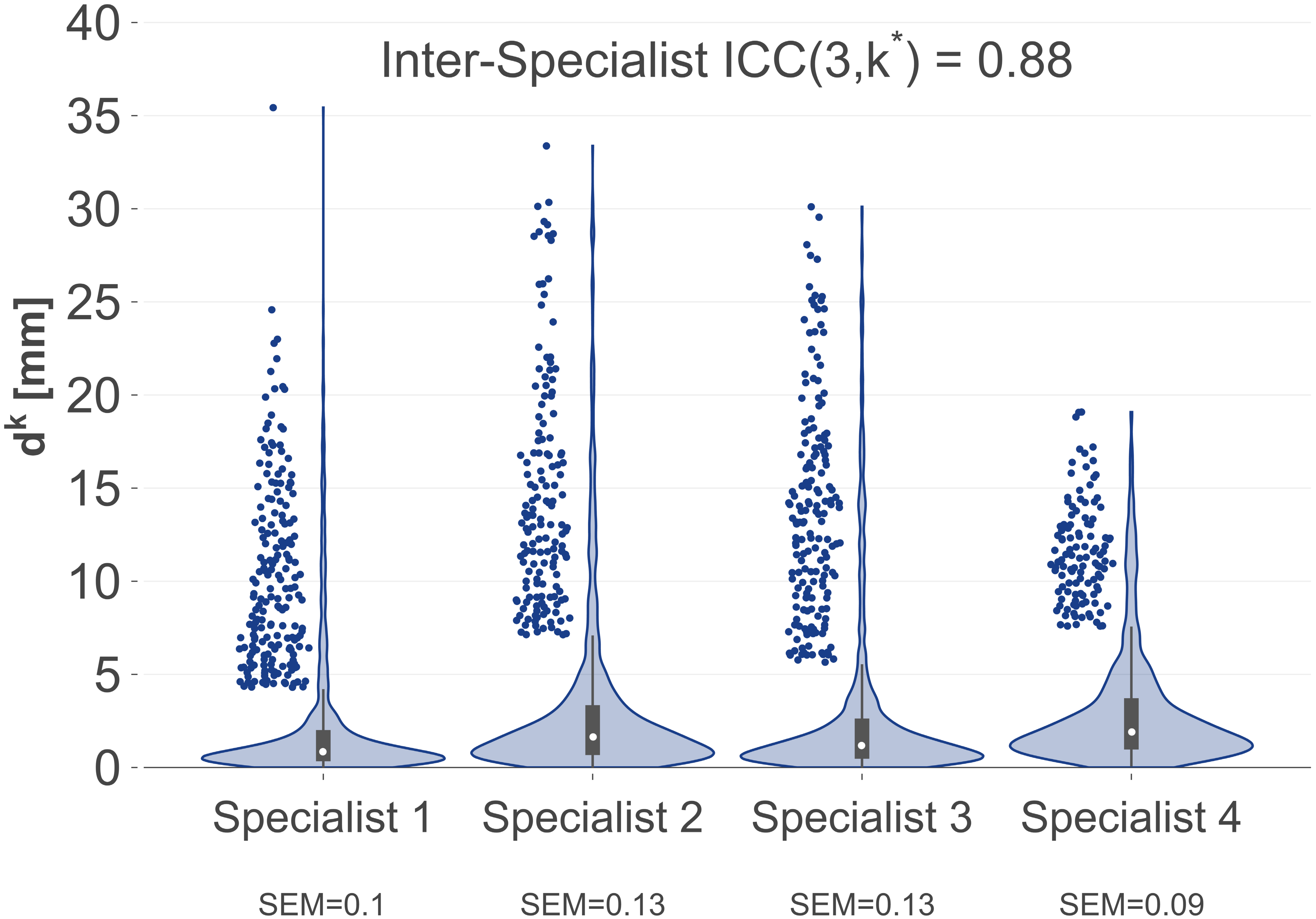}
  \caption{Violin plots show the labeling performance of specialists on the test-set. We plot Euclidean distances $\boldsymbol{d}^{k}$ between the Specialist label positions $\boldsymbol{P}^{k}_{x,y}$ and the mean label positions of the remaining three specialists $\boldsymbol{P}^{j}_{x,y}$. Violin widths are normalized to the total number of included labels in the test-set. Boxplots and whiskers are indicated in gray color. The median is depicted as white circle inside the box. Outliers are shown as filled dots. The interclass correlation coefficient (ICC) at the top is based on a mean-rating ($k^{*} = 3$) and absolute-agreement in a 2-way mixed-effects model. SEMs for individual specialists are shown horizontally underneath the graphs.}
  \label{fig:InterHumanExpert}
\end{figure}

\section{RESULTS AND DISCUSSION}
\label{sec:results}

In this section, we first investigate inter-rate reliability's and than compare our model predictions to the reference labels. Moreover, we assess performance of other recent, and open-source, available MTJ trackers on our test-set and compare them with our model results. In Table \ref{tab:Results} an overview of model results is presented. Results are obtained from the full test-set, which includes data from all considered instruments (Aixplorer, Esaote and Telemed), two muscles (MG and LG) and two movements (MVC and PT). Among the considered automated methods, our approach shows the best performance both in terms of RMSE (RMSE\textsuperscript{M} $= 4.89$ mm) and SEM (SEM\textsuperscript{M} $= 0.10$ mm). 

\subsection*{Evaluation of specialists}
Violin plots in Fig. \ref{fig:InterHumanExpert} show labeling results for each specialist. We plot Euclidean distances $\boldsymbol{d}^{k}$ between the Specialist label positions $\boldsymbol{P}^{k}_{x,y}$ and the  mean label positions of the remaining three specialists $\boldsymbol{P}^{j}_{x,y}$. Median and SEM values are comparable between the specialists. Few larger errors of up to 35.43 mm were found for annotations of specialists 1-3. The inter-rater assessment using ICC confirms good inter-rater reliability for specialists (ICC = 0.88, 95\% confidence interval = [0.87, 0.89]).

\subsection*{Comparison to specialists}
Table \ref{tab:Results} shows that our trained network (RMSE\textsuperscript{M}$ = 4.89$ mm) is within the range of the performance level of specialists (RMSE\textsuperscript{S}$ = 5.05 \pm0.62$ mm) on the test-set. The SEM suggests that our model produces consistent results, comparable to the individual specialists. Furthermore, it takes 107.4 s in total (tested on a NVIDIA GeForce RTX2070 graphical processing unit (GPU)) to predict all MTJ positions of the test-set using our algorithm whereas individual specialists need on average $2.76 \pm1.11$ hours to label the same dataset. This shows that our method can accelerate data analyses times by a factor of 100 on a low cost GPU.
\begin{figure}[ht!]
  \centering
  \vspace{-0.3cm}
  \includegraphics[width=0.45\textwidth, clip, trim={0 0 0 0}]{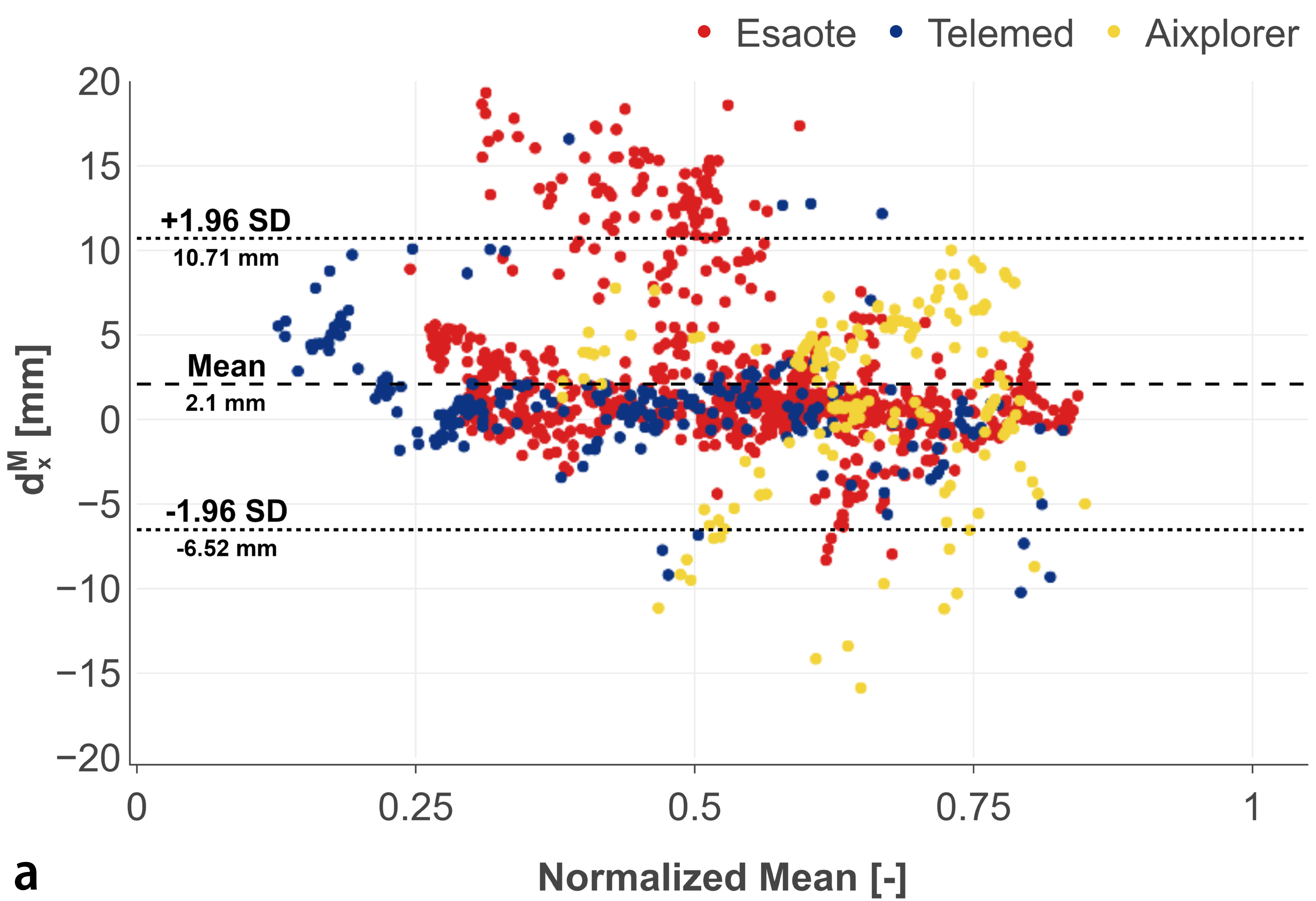}
  
   \vspace{0.3cm}
  
  \includegraphics[width=0.45\textwidth, clip, trim={0 0 0 0}]{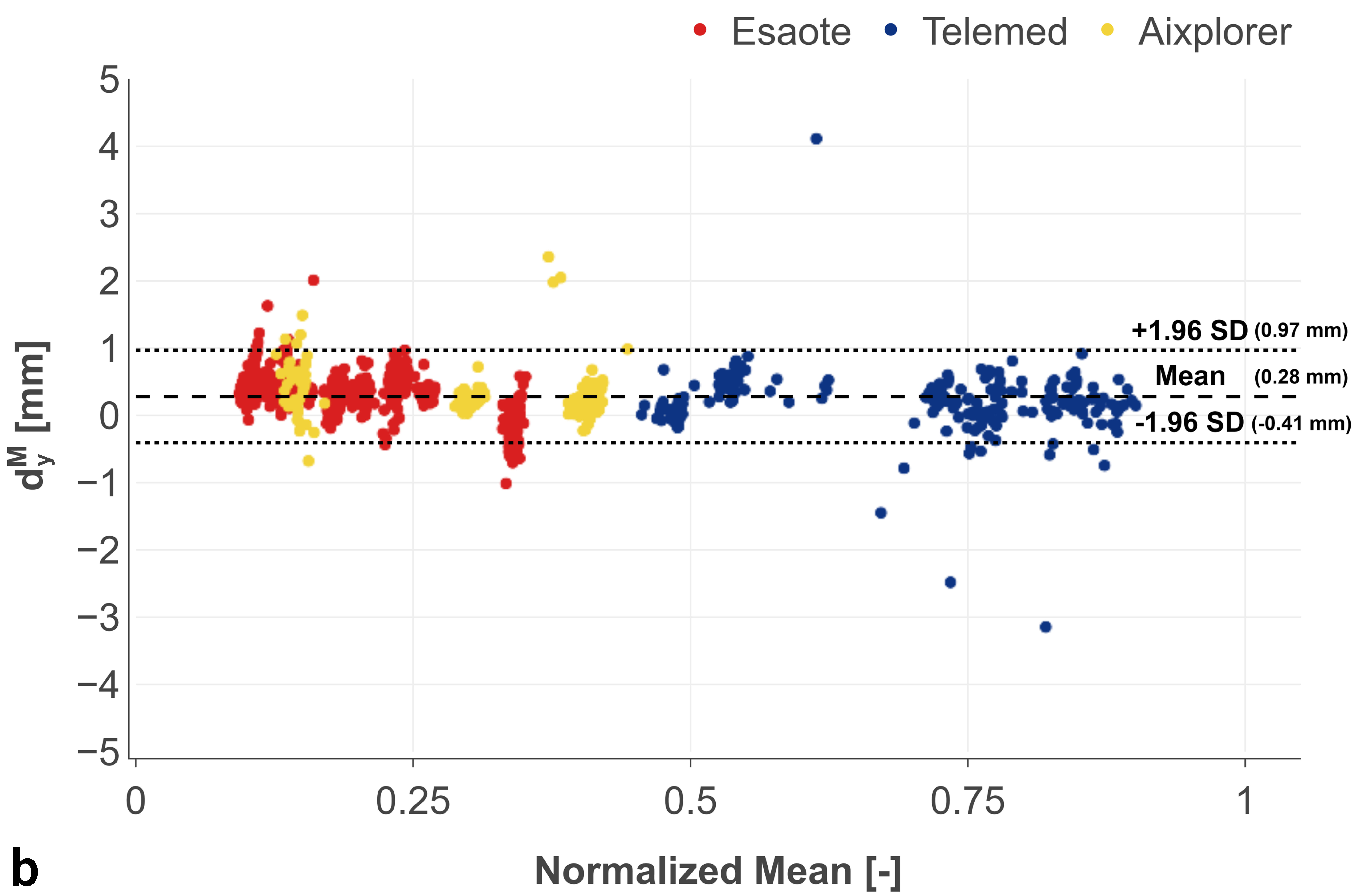}
  \caption{Bland–Altman plots for inter-rater agreement analysis (1344 test-set samples). In figure \textbf{a} and \textbf{b} we show results for MTJ x- and y-coordinates, respectively. Comparisons are illustrated by differences between pairs of measurements (${d}^{M}_{x}$, ${d}^{M}_{y}$) as a function of the normalized mean measures (normalization to instrument image size). Dashed and dotted lines depict bias and 95\% limits of agreement, respectively. The color facets show the distribution of instruments: Esaote - red, Telemed - blue, Aixplorer - yellow.}
  \label{fig:ba}
\end{figure}
\begin{figure}[h!]
  \centering
  \includegraphics[width=0.43\textwidth, clip, trim={0 0 0 0}]{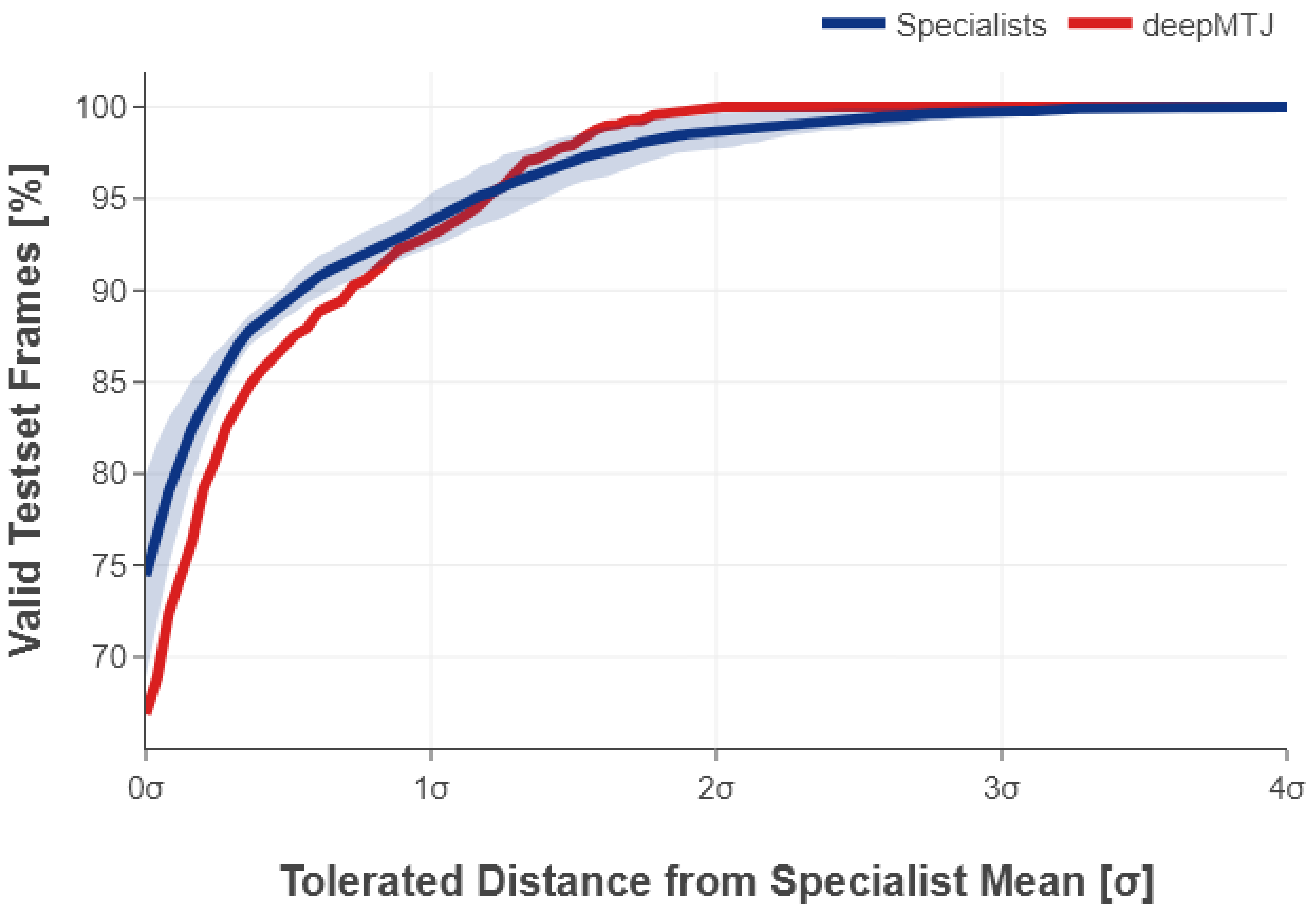}
  \caption{Line plots (our model in red color and specialists in blue color) show valid numbers of frames in percent (y-axis) with increasing tolerance distance (x-axis). Tolerance distance is depicted as a multiple $n^{*}$ of absolute standard deviation $\bar{\sigma}^{S}$ from reference label positions. $\bar{\boldsymbol{P}}^{S}_{x,y}$.}
  \label{fig:toleranceRd}
  \vspace{-0.3cm}
\end{figure}

Bland-Altman plots in Fig. \ref{fig:ba} show that our model
has no substantial bias in x- (2.1 mm) and y-direction (0.28 mm) of the detected MTJ position. Larger deviations in the x-coordinate (Fig. \ref{fig:ba} a) result from the leading travel direction (Fig. \ref{fig:MTJ} b) of the MTJ. We assume that image quality (i.e., CNR) influences the estimation of MTJ positions for Esaote and Aixplorer, causing larger scattering in x direction. None of the measured biases were found to be proportional to averaged values, indicating that manual and automatic analyses agree equally through the range of measurements.

In Fig. \ref{fig:toleranceRd}, we plot the percentage of correct MTJ detections and correct specialist labels as function of tolerance distance. Tolerance distance is depicted as a multiple $n^{*}$ of absolute standard deviation $\bar{\sigma}^{S}$ from reference label positions $\bar{\boldsymbol{P}}^{S}_{x,y}$. Results show that the number of correct samples is similar among our model and specialists. A slightly higher performance is achieved by specialists for small $n^{*}$. The neural network appears to be more robust in general, as can be seen from the close to 100\% correct detections with a tolerance distance $>2\cdot\bar{\sigma}^{S}$. In other words, the network shows no random false detections, that can occur during manual labeling (e.g., due to attention loss in monotonous tasks \cite{b:Robertson2012}).

\begin{figure}[t!]
  \centering

  \includegraphics[width=0.45\textwidth, clip, trim={0 0 0 0}]{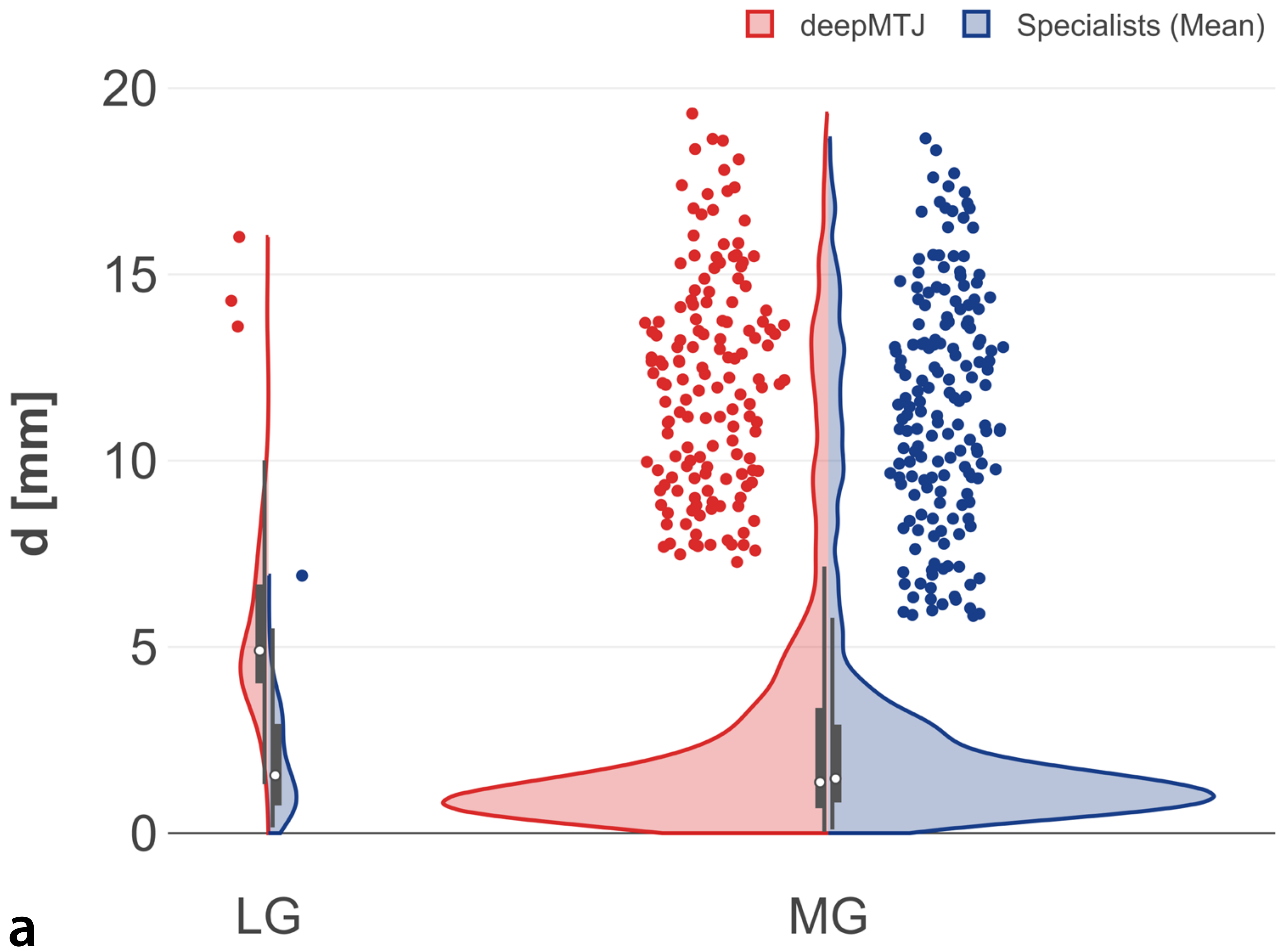}
  
  \vspace{0.3cm}
  
  \includegraphics[width=0.45\textwidth, clip, trim={0 0 0 0}]{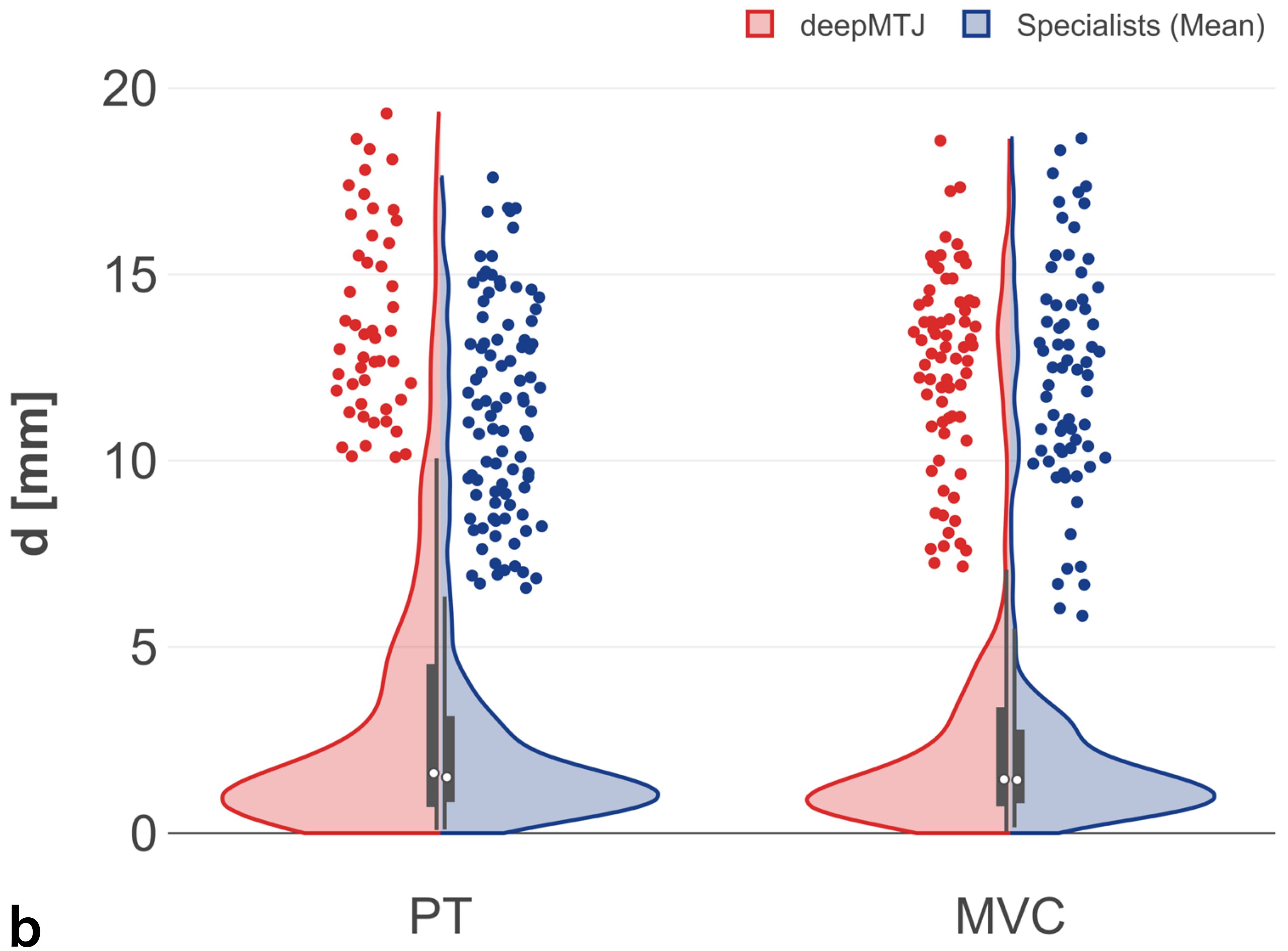}
  \caption{Breakdown of our model results and specialist deviations in terms of \textbf{a} muscles and \textbf{b} movements. We plot Euclidean distances $\boldsymbol{d}^{M}$ between our model label positions $\boldsymbol{P}^{M}_{x,y}$ and reference label positions $\bar{\boldsymbol{P}}^{S}_{x,y}$ (red half-violin, negative side) as well as mean specialist deviations $\bar{\boldsymbol{d}}^{S}$ (blue half-violin, positive side). Violin widths are normalized to the total number of included labels in the test-set. Boxplots and whiskers are indicated in gray color. The median is depicted as white circle inside the box. Outliers are shown as filled, jittered dots.} 
  \label{fig:inst_muscle_move}
\end{figure}
%
%
%
%
\setlength{\tabcolsep}{0.32em}
\begin{table}[t!]
\caption{Functional assessment of the dataset}
\label{tab:results_MuMov}
    \vspace{-0.2cm} 
    \renewcommand{\arraystretch}{1.20}
    \begin{center}
    \begin{tabular}{ccc}
        \toprule
        \multirow{2}{*}{\textbf{Category}}
        &\textbf{RMSE$^{M}$}
        &\textbf{RMSE$^{S}$}\\
        &[mm]
        &[mm]\\
        \midrule
        \rule{0pt}{12pt}LG
        &6.03
        &$2.26\pm{0.14}$\\
        \rule{0pt}{12pt}MG
        &4.79
        &$5.15\pm{0.65}$\\
        \rule{0pt}{12pt}PT
        &5.02
        &$5.15\pm{1.14}$\\
        \rule{0pt}{12pt}MVC
        &4.74
        &$4.47\pm{0.46}$\\
        \bottomrule
    \end{tabular}
    \end{center}
        \begin{tablenotes}
        \item RMSE$^{M}$ ... root-mean-square-error of our model, RMSE$^{S}$ ... root-mean-square-error of specialists, LG ... lateral gastrocnemius, MG ... medial gastrocnemius, MVC ... isometric maximum voluntary contraction, PT ... passive torque movement.\\
    \end{tablenotes}
\vspace{-0.3cm}
\end{table}
\subsection*{Performance on different muscles and movements}
We decompose our test-set with respect to included muscles and movements, and then compare results of our algorithm $\boldsymbol{d}^{M}$ to mean deviations of specialists $\bar{\boldsymbol{d}}^{S}$ for each category. In Figure \ref{fig:inst_muscle_move}\textbf{a}, we show performances on images of LG and MG muscle-tendon units. In Figure \ref{fig:inst_muscle_move}\textbf{b}, we highlight all video frames for our considered movements (MVC and PT). RMSEs shown in Table \ref{tab:results_MuMov} and kernel densities in Fig. \ref{fig:inst_muscle_move}\textbf{a} and \ref{fig:inst_muscle_move}\textbf{b} are in good agreement with larger deviations for predictions on the LG. In case of the LG, the share of labeled ground truth data in comparison to the total data volume was small (9.4\%). In addition, LG data of the test-set was recorded by the US system (Aixplorer) with the least amount of training data compared to other included devices. Therefore, we assume that performance shortcomings stem from the low training and test dataset volume. Furthermore, varying morphologies of gastrocnemii heads around the muscle tendon junction could also influence prediction qualities of our model. In terms of movements, the two dataset sizes were balanced. The results show that our model has no preference for a specific movement. Comparisons with manual tracking suggests that MTJs, during controlled passive and active contractions, can be reliably and accurately tracked using our model.

\subsection*{Comparison with other MTJ trackers}
We evaluated the performance of recently proposed MTJ tracking approaches on our test-set (Table \ref{tab:Results}) and found that neither could reach the level of specialists. In particular, the machine learning approaches were trained on machine specific data collected with US systems of one vendor. The approach by Leitner \textit{et al.}\cite{c:LeitnerJarolim2020} was trained on Esaote data only, while the method by Krupenevich \textit{et al.}\cite{j:Krupenevich2021} is based entirely on Telemed data. 
When evaluating these algorithms on our test-set data from the respective machine type, the methods by Leitner \textit{et al.} \cite{c:LeitnerJarolim2020} and Krupenevich \textit{et al.} \cite{j:Krupenevich2021} achieve RMSEs of 5.57 mm and 12.06 mm on Esaote and Telemed data, respectively. Therefore, we conclude that these two machine-learning approaches are domain specific and provide poor performance on our diverse test-set. We associate the lack in performance with the lack of data diversity during training. Hence, it seems that the diversity of instruments, inter-laboratory data and inter-observer settings, as represented in our test-set impacts prediction qualities of previously proposed algorithms.

\begin{figure}[ht!]
  \centering
  \vspace{-0.3cm}
  \includegraphics[width=0.45\textwidth, clip, trim={0 0 0 0}]{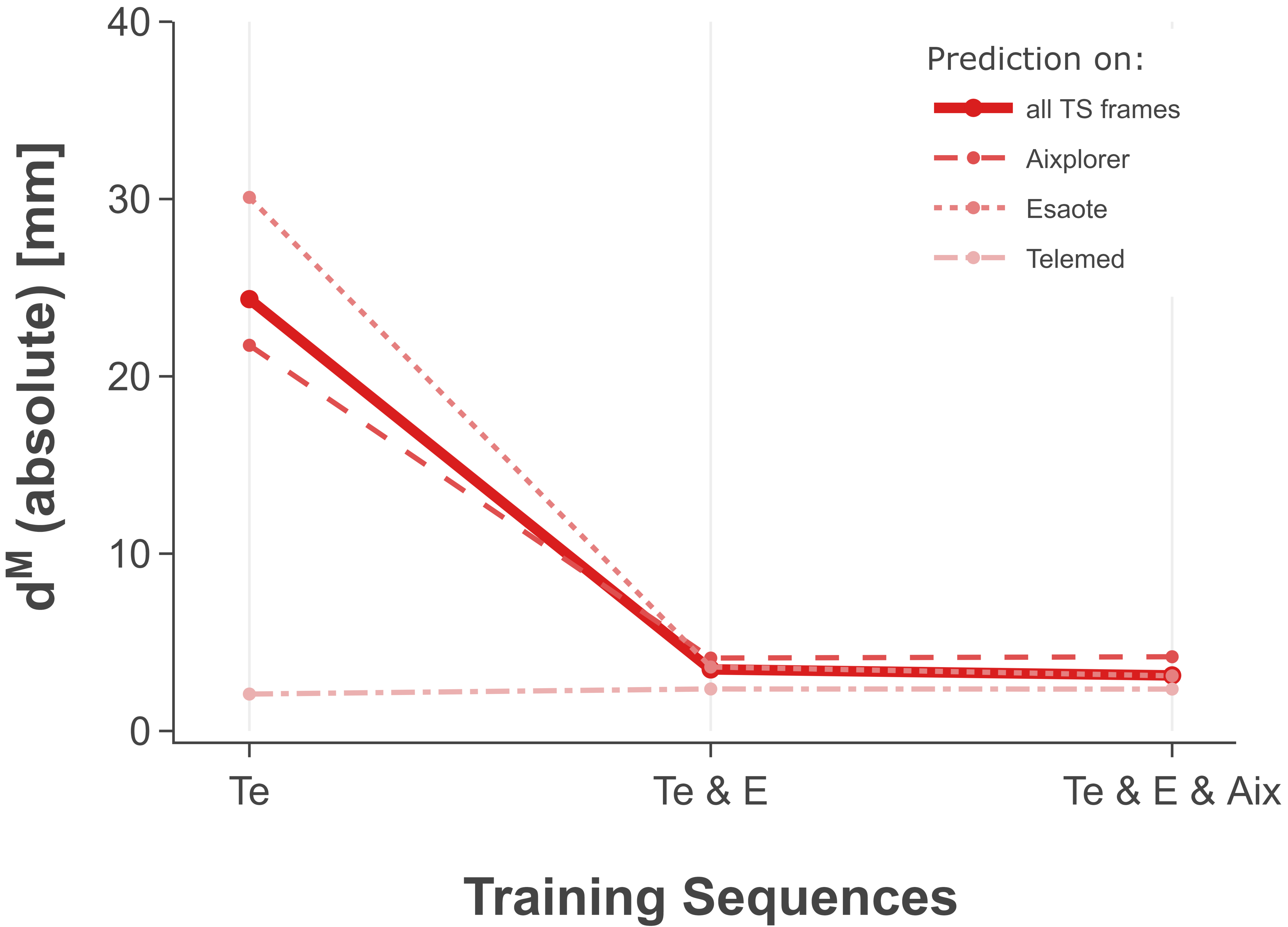}
  \caption{The line plots show absolute Euclidean distances $\boldsymbol{d}^{M}$ (y-axis) between our model label positions $\boldsymbol{P}^{M}_{x,y}$ and reference label positions $\bar{\boldsymbol{P}}^{S}_{x,y}$ for the full test-set (solid line), Aixplorer test-set data (dashed line), Esaote test-set data (dash-dotted line) and Telemed test-set data (dotted line). On the x-axis we denoted the training sequence: 1. training on Telemed (Te), 2. training on Telemed and Esaote (Te \& E) and 3. training on our final model with Telemed, Esaote and Aixplorer (Te \& E \& Aix)}
  \label{fig:training_sequence}
\end{figure}

\subsection*{Domain Generalization}
\label{sec:generalization}
We observe that image qualities differ across US machines of different vendors (Fig. \ref{fig:MTJ}). This can originate in the US image formation (beamforming), filtering and other signal processing techniques (e.g. time-gain compensation) implemented on US systems themselves or preset by operators. Image appearances may also vary due to different applications of US systems in experimental setups. Leitner \textit{et al.} \cite{j:Leitner2019} have shown that, for example, the Telemed system is widely used to collect data during movement with increased frame-rates at the cost of decreasing image qualities. Other devices, such as the Esaote system, offer transducers with wide probe heads to collect data on movements where larger scale length changes occur, decreasing frame-rates. To counteract sensitivity for noise coming from experimental setups and applications we have added data augmentation methods (as
depicted in Sect. II-B). Furthermore, the evaluation of methods by Leitner \textit{et al.} \cite{c:LeitnerJarolim2020} and Krupenevich \textit{et al.} \cite{j:Krupenevich2021} on our multi-instrumental test-set has demonstrated that the performance of a neural network strongly depends on instrument types used for training and testing (Table \ref{tab:Results}). Therefore, our model was trained with US images from three of the four most common US instruments in biomechanical research \cite{j:Leitner2019, j:Hooren2020}. 
%
%
%
%
\setlength{\tabcolsep}{0.32em}
\begin{table}[!ht]
\caption{Distribution of errors on instrumental data during sequential training steps}
\label{tab:results_Inst}
    \vspace{-0.2cm} 
    \renewcommand{\arraystretch}{1.20}
    \begin{center}
    \begin{tabular}{cccc}
        \toprule
        \multirow{2}{*}{\textbf{Training Sequence}}
        &\multicolumn{3}{c}{\textbf{RMSE$^{M\ddagger}$ [mm]}}\\
        &Aixplorer
        &Esaote
        &Telemed\\ 
        \midrule
        \rule{0pt}{12pt}\{Telemed\}
        &41.32
        &48.66
        &3.32\\
        \rule{0pt}{12pt}\{Telemed, Esaote\}
        &5.11
        &6.31
        &4.43\\
        \rule{0pt}{12pt}\{Telemed, Esaote, Aixplorer\}
        &5.17
        &5.09
        &3.60\\
        \bottomrule
    \end{tabular}
    \end{center}
        \begin{tablenotes}
        \item $^\ddagger$ RMSEs are calculated for data of particular instruments during each training sequence.\\
    \end{tablenotes}
\vspace{-0.3cm}
\end{table}

In order to analyse the generalization of our network to novel instrumental data, we sequentially introduced datasets from the individual instruments during our model training (Sect. \ref{sec:methods.train}). Fig. \ref{fig:training_sequence} shows the mean absolute deviations of the two intermediate models (trained with images from \{Telemed\} and \{Telemed, Esaote\}) and the final model (trained with images from all instruments \{Telemed, Esaote, Aixplorer\}). For the model that was trained with Telemed data only, we can observe excellent performance for the Telemed images, but the model fails for the other two machine types, similar to the studies by Leitner \textit{et al.} \cite{c:LeitnerJarolim2020} and Krupenevich \textit{et al.} \cite{j:Krupenevich2021}.

By adding the Esaote data to the model training, we obtain a significant performance increase for Esaote frames as well as for Aixplorer samples (see Table \ref{tab:results_Inst}). The model trained with solely Telemed and Esaote images, achieves a high performance for Aixplorer data (RMSE of 5.11), although it was not included in the training set. This suggests a successful generalization to MTJ tracking in US imagery. Thus, we expect that the final training stage of our model \{Telemed, Esaote, Aixplorer\} provides a similar performance on novel US instrumental data. Moreover, results in Table \ref{tab:results_Inst} show that including images from multiple instruments in the model training does not decrease performance for the individual instruments. Hence, models trained with images from \{Telemed\}, \{Telemed, Esaote\} or \{Telemed, Esaote, Aixplorer\} achieve similar performances on the Telemed test-set.

From our multi-instrument training approach and the applied data augmentation, we expect that our model can be applied to a larger variety of different experimental setups (e.g., instrument types, transducer rotations and shifts, image distortions due to movement,...) with less susceptibility to errors.

\section{CONCLUSION}
\label{sec:conclusion}

In this work, we presented a data-driven deep-learning method for the estimation of MTJ positions in US images. Our approach is based on a CNN trained to infer MTJ positions across a variety of US systems from different vendors, collected in independent laboratories from diverse observers, on distinct muscles and movements. We used data augmentation methods and soft labeling to counteract label noise introduced by the experimental setups and inter-observer variability, respectively. A diverse test-set was created and labeled by four independent specialists, in order to provide more objective estimates of MTJ positions. Our method was able to identifiy MTJ positions in 99.2\% of cases, with a root-mean-square deviation of 4.89 mm. 
This error is in the same range as the variation of human specialists (RMSE\textsuperscript{S}$ = 5.05 \pm0.62$ mm). Therefore, our method provides human-like performance on a divers test-set and requires only a fraction of manual labeling times (approx. 100 times faster). We demonstrated that our approach is applicable to generalize to data collected on different US machine types. We made all our codes, trained models and test-set (1344 labeled ultrasound images) publicly available and provide our model as a free-to-use online service under https://deepmtj.org/.

\appendices
\section*{Acknowledgment}
C. Leitner would like to acknowledge Martin Sust (University of Graz) for the support of his research.

\section{}
\label{sec:Appendix}
\begin{figure*}[hbt!]
  \centering
  \includegraphics[width=0.9\textwidth, clip, trim={0 0 0 0}]{{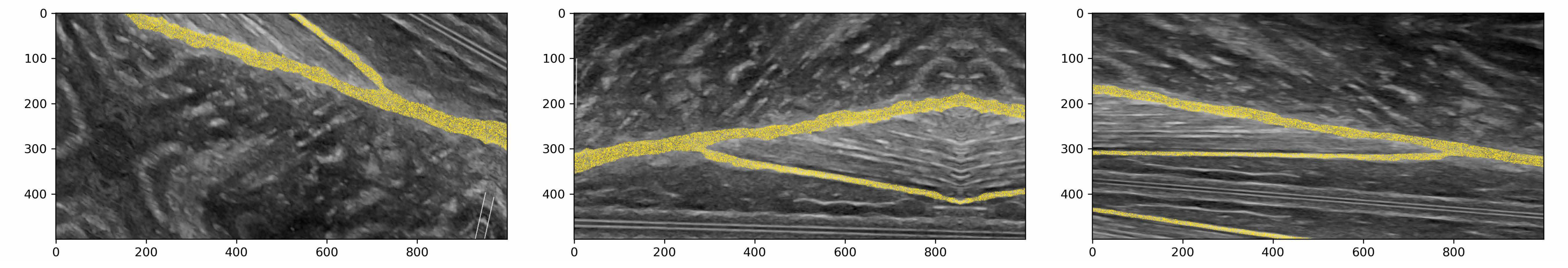}}
  \caption{Figure shows three augmented frames of the same image. We used the \texttt{ImageDataGenerator} implemented in TensorFlow \cite{w:tensorflow2015} to additionally generate random data with the following properties: \texttt{rotation-range=20, horizontal-flip=True, vertical-flip=True, zoom-range=0.3, width-shift-range=0.1, height-shift-range=0.1, shear-range=0.2, fill-mode=reflect}.}
  \label{fig:Augmentation}
\end{figure*}

\begin{figure*}[hbt!]
  \centering
  \includegraphics[width=0.9\textwidth, clip, trim={0 0 0 0}]{{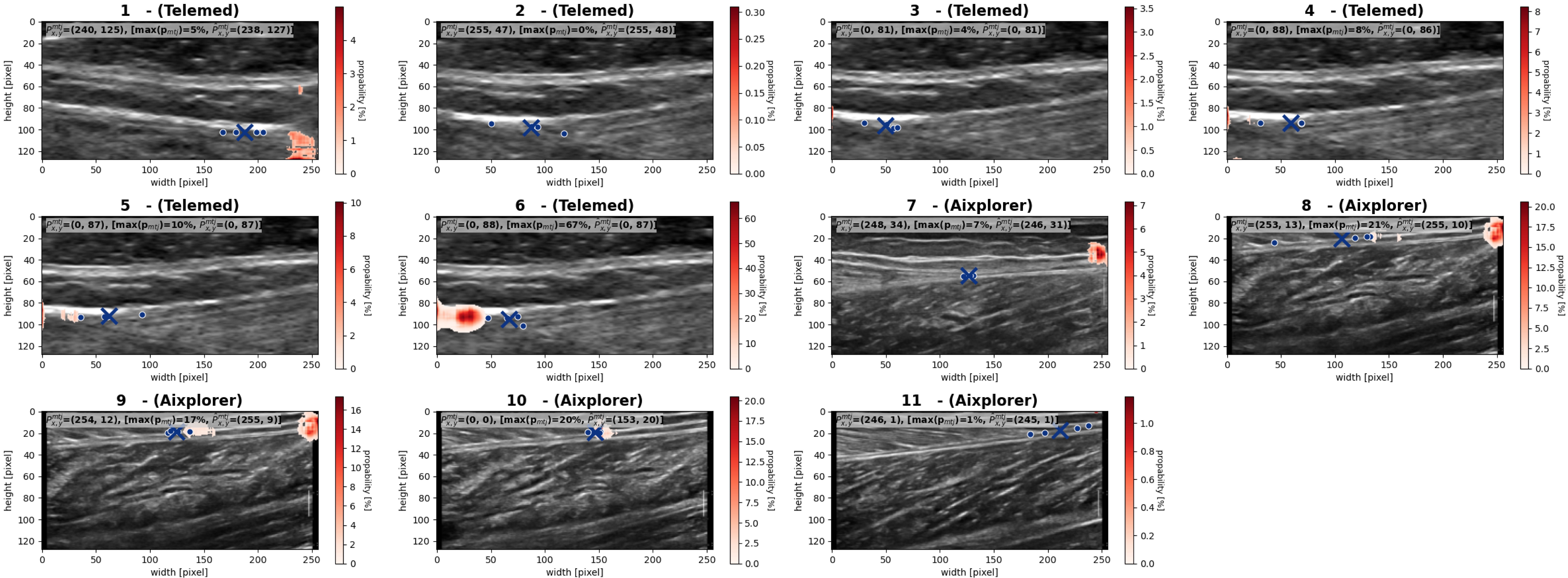}}
  \caption{Excluded images for error-case 1 and 2. Blue crosses indicate pixel positions of reference labels. MTJ probability densities generated by our model are given in shades of red color.}
  \label{fig:EC12}
\end{figure*}

\begin{figure*}[hbt!]
  \centering
   \includegraphics[width=0.9\textwidth, clip, trim={0 0 0 0}]{{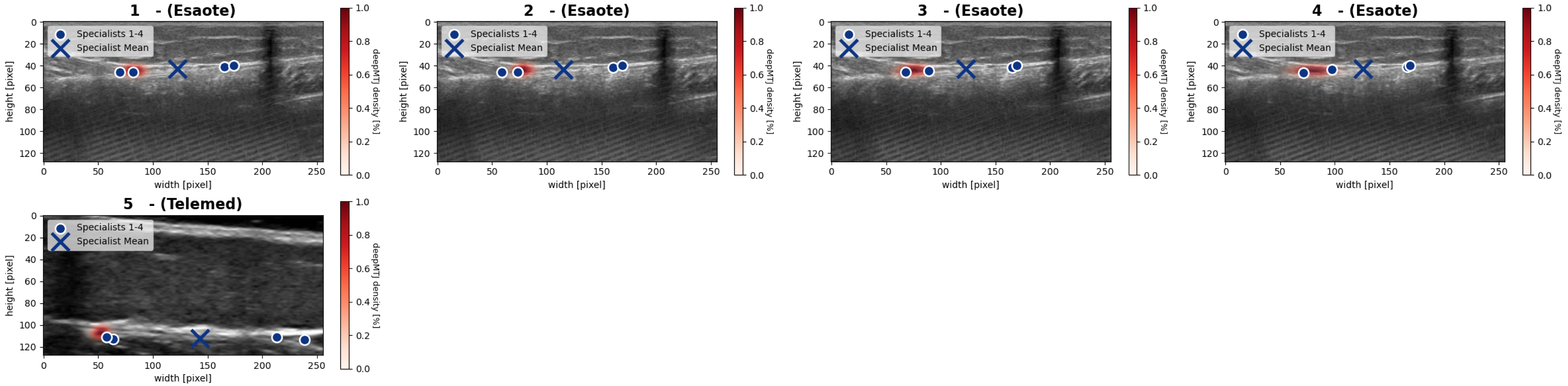}}
  \caption{Excluded images for error-case 3. Blue crosses indicate pixel positions of reference labels. MTJ positions chosen by individual specialists are depicted as blue points with white edges. Gaussian density distributions for MTJ positions generated by our model are given in red color shading.}
  \label{fig:EC3}
\vspace{-0.5cm}
\end{figure*}

In this section we provide additional figures supporting our work. Fig. \ref{fig:Augmentation} showcases three augmented frames of the same image (MTJ collected on an Esaote system) using our settings for data augmentation. In Fig. \ref{fig:EC12} and Fig. \ref{fig:EC3}, we show identified and excluded images from error-cases 1 and 2 as well as error-case 3, respectively.

\bibliography{references.bib}

\begin{thebibliography}{10}
\providecommand{\url}[1]{#1}
\csname url@samestyle\endcsname
\providecommand{\newblock}{\relax}
\providecommand{\bibinfo}[2]{#2}
\providecommand{\BIBentrySTDinterwordspacing}{\spaceskip=0pt\relax}
\providecommand{\BIBentryALTinterwordstretchfactor}{4}
\providecommand{\BIBentryALTinterwordspacing}{\spaceskip=\fontdimen2\font plus
\BIBentryALTinterwordstretchfactor\fontdimen3\font minus
  \fontdimen4\font\relax}
\providecommand{\BIBforeignlanguage}[2]{{%
\expandafter\ifx\csname l@#1\endcsname\relax
\typeout{** WARNING: IEEEtran.bst: No hyphenation pattern has been}%
\typeout{** loaded for the language `#1'. Using the pattern for}%
\typeout{** the default language instead.}%
\else
\language=\csname l@#1\endcsname
\fi
#2}}
\providecommand{\BIBdecl}{\relax}
\BIBdecl

\bibitem{j:Konrad2014:1}
A.~Konrad and M.~Tilp, ``{Increased range of motion after static stretching is
  not due to changes in muscle and tendon structures},'' \emph{Clinical
  Biomechanics}, vol.~29, no.~6, pp. 636--642, 2014.

\bibitem{j:Konrad2015}
A.~Konrad \emph{et~al.}, ``{Effect of PNF stretching training on the properties
  of human muscle and tendon structures},'' \emph{Scandinavian Journal of
  Medicine and Science in Sports}, vol.~25, no.~3, pp. 346--355, 6 2015.

\bibitem{j:Konrad2014:2}
A.~Konrad and M.~Tilp, ``{Effects of ballistic streching training on the
  properties of human muscle and tendon structures},'' \emph{Journal of Applied
  Physiology}, vol. 117, no.~1, pp. 29--35, 7 2014.

\bibitem{j:Kruse2017}
A.~Kruse \emph{et~al.}, ``{Mechanical muscle and tendon properties of the
  plantar flexors are altered even in highly functional children with spastic
  cerebral palsy},'' \emph{Clinical Biomechanics}, vol.~50, pp. 139--144, 12
  2017.

\bibitem{j:Kruse2018}
A.~Kruse \emph{et~al.}, ``{Muscle and tendon morphology alterations in children
  and adolescents with mild forms of spastic cerebral palsy},'' \emph{BMC
  Pediatrics}, vol.~18, 12 2018.

\bibitem{j:Kruse2019}
A.~Kruse \emph{et~al.}, ``{The effect of functional home-based strength
  training programs on the mechano-morphological properties of the plantar
  flexor muscle-tendon unit in children with spastic cerebral palsy},''
  \emph{Pediatric Exercise Science}, vol.~31, no.~1, pp. 67--75, 2 2019.

\bibitem{j:Komi2000}
P.~V. Komi, ``{Stretch-shortening cycle: a powerful model to study normal and
  fatigued muscle},'' \emph{Journal of Biomechanics}, vol.~33, no.~10, pp.
  1197--1206, 10 2000.

\bibitem{j:Alexander1991}
R.~M. Alexander, ``{Energy-saving mechanisms in walking and running},''
  \emph{Journal of Experimental Biology}, vol. 160, no.~1, pp. 55--69, 1991.

\bibitem{j:Lichtwark2006}
G.~A. Lichtwark and A.~M. Wilson, ``{Interactions between the human
  gastrocnemius muscle and the Achilles tendon during incline, level and
  decline locomotion},'' \emph{Journal of Experimental Biology}, vol. 209,
  no.~21, pp. 4379--4388, 11 2006.

\bibitem{j:Magnusson2008}
S.~P. Magnusson \emph{et~al.}, ``\BIBforeignlanguage{eng}{{Human tendon
  behaviour and adaptation, in vivo.}}'' \emph{\BIBforeignlanguage{eng}{The
  Journal of physiology}}, vol. 586, no.~1, pp. 71--81, 1 2008.

\bibitem{j:Roberts1997}
T.~J. Roberts \emph{et~al.}, ``{Muscular force in running turkeys: the economy
  of minimizing work.}'' \emph{Science (New York, N.Y.)}, vol. 275, no. 5303,
  pp. 1113--1115, 2 1997.

\bibitem{j:Arampatzis2020}
A.~Arampatzis \emph{et~al.}, ``{Individualized Muscle-Tendon Assessment and
  Training},'' \emph{Frontiers in Physiology}, vol.~11, 6 2020.

\bibitem{j:Hoesl2020}
M.~H{\"{o}}sl \emph{et~al.}, ``{Impact of Altered Gastrocnemius Morphometrics
  and Fascicle Behavior on Walking Patterns in Children With Spastic Cerebral
  Palsy},'' \emph{Frontiers in Physiology | www.frontiersin.org}, vol.~11, p.
  518134, 2020.

\bibitem{j:Lee2008}
S.~Lee \emph{et~al.}, ``{An algorithm for automated analysis of ultrasound
  images to measure tendon excursion in vivo},'' \emph{Journal of Applied
  Biomechanics}, vol.~24, no.~1, pp. 75--82, 2008.

\bibitem{j:Zhou2018}
G.-Q. Zhou \emph{et~al.}, ``{Automatic Myotendinous Junction Tracking in
  Ultrasound Images with Phase-Based Segmentation},'' \emph{BioMed Research
  International}, vol. 2018, pp. 1--12, 2018.

\bibitem{j:Cenni2019}
F.~Cenni \emph{et~al.}, ``{Semi-automatic methods for tracking the medial
  gastrocnemius muscle–tendon junction using ultrasound: a validation
  study},'' \emph{Experimental Physiology}, vol. 105, no.~1, pp. 120--131,
  2020.

\bibitem{c:LeitnerJarolim2020}
C.~Leitner \emph{et~al.}, ``Automatic tracking of the muscle tendon junction in
  healthy and impaired subjects using deep learning,'' in \emph{in Proc. 42nd
  Conferences of the IEEE Engineering in Medicine and Biology Society}.\hskip
  1em plus 0.5em minus 0.4em\relax IEEE, 07 2020.

\bibitem{j:Krupenevich2021}
R.~L. Krupenevich \emph{et~al.}, ``{Automated Analysis of Medial Gastrocnemius
  Muscle-Tendon Junction Displacements in Heathy Young Adults During Isolated
  Contractions and Walking Using Deep Neural Networks},'' \emph{Computer
  Methods and Programs in Biomedicine}, p. 106120, 4 2021.

\bibitem{j:Charvet2012}
B.~Charvet \emph{et~al.}, ``{The development of the myotendinous junction. A
  review},'' \emph{Ligaments and Tendons Journal}, vol.~2, no.~2, pp. 53--63,
  2012.

\bibitem{j:Leitner2019}
C.~Leitner \emph{et~al.}, ``{Ultrasound as a Tool to Study Muscle–Tendon
  Functions during Locomotion: A Systematic Review of Applications},''
  \emph{Sensors}, vol.~19, no.~19, p. 4316, 10 2019.

\bibitem{j:Werkhausen2018}
A.~Werkhausen \emph{et~al.}, ``{Effect of training-induced changes in Achilles
  tendon stiffness on muscle-tendon behavior during landing},'' \emph{Frontiers
  in Physiology}, vol.~9, no. JUN, 6 2018.

\bibitem{j:Barber2017}
L.~Barber \emph{et~al.}, ``{Medial gastrocnemius and soleus muscle-tendon unit,
  fascicle, and tendon interaction during walking in children with cerebral
  palsy},'' \emph{Developmental Medicine and Child Neurology}, vol.~59, no.~8,
  pp. 843--851, 8 2017.

\bibitem{j:Ohrndorf2010}
S.~Ohrndorf \emph{et~al.}, ``{Is musculoskeletal ultrasonography an
  operator-dependent method or a fast and reliably teachable diagnostic tool?
  Interreader agreements of three ultrasonographers with different training
  levels},'' \emph{International Journal of Rheumatology}, vol. 2010, 2010.

\bibitem{j:Kharazi2020}
M.~Kharazi \emph{et~al.}, ``{Quantifying mechanical loading and elastic strain
  energy of the human Achilles tendon 1 during walking and running},''
  \emph{bioRxiv}, p. 2020.09.05.284182, 9 2020.

\bibitem{j:Hooren2020}
X.~Bas, V.~Hooren \emph{et~al.}, ``{Ultrasound imaging to assess skeletal
  muscle architecture during movements: a systematic review of methods,
  reliability, and challenges},'' \emph{J Appl Physiol}, vol. 128, pp.
  978--999, 2020.

\bibitem{proc:Sukhwan2001}
L.~Sukhwan and A.~El~Gamal, ``{Optical flow estimation using high frame rate
  sequences},'' in \emph{Proceedings 2001 International Conference on Image
  Processing}.\hskip 1em plus 0.5em minus 0.4em\relax Thessaloniki, Greece:
  IEEE, 2001.

\bibitem{j:Ng2011}
A.~Ng and J.~Swanevelder, ``{Resolution in ultrasound imaging},''
  \emph{Continuing Education in Anaesthesia, Critical Care and Pain}, vol.~11,
  no.~5, pp. 186--192, 10 2011.

\bibitem{j:Cronin2020}
N.~J. Cronin \emph{et~al.}, ``{Fully automated analysis of muscle architecture
  from B-mode ultrasound images with deep learning},'' \emph{arxiv}, 9 2020.

\bibitem{Goodfellow-et-al-2016}
I.~Goodfellow \emph{et~al.}, \emph{Deep Learning}.\hskip 1em plus 0.5em minus
  0.4em\relax MIT Press, 2016, \url{http://www.deeplearningbook.org}.

\bibitem{lecun2015deep}
Y.~LeCun \emph{et~al.}, ``Deep learning,'' \emph{nature}, vol. 521, no. 7553,
  pp. 436--444, 2015.

\bibitem{c:Englmair2020}
B.~Englmair \emph{et~al.}, ``{Improved Tracking of Muscle Tendon Junctions in
  Ultrasound Images Using Speckle Reduction},'' in \emph{Studies in health
  technology and informatics}, vol. 271.\hskip 1em plus 0.5em minus 0.4em\relax
  NLM (Medline), 6 2020, pp. 1--8.

\bibitem{c:He2016}
K.~He, X.~Zhang, S.~Ren, and J.~Sun, ``{Deep residual learning for image
  recognition},'' in \emph{Proceedings of the IEEE Computer Society Conference
  on Computer Vision and Pattern Recognition}, vol. 2016-December.\hskip 1em
  plus 0.5em minus 0.4em\relax IEEE Computer Society, 12 2016, pp. 770--778.

\bibitem{j:Jetley2018}
S.~Jetley \emph{et~al.}, ``{Learn To Pay Attention},'' \emph{arxiv}, 4 2018.

\bibitem{c:Sandler2018}
M.~Sandler \emph{et~al.}, ``{MobileNetV2: Inverted Residuals and Linear
  Bottlenecks},'' in \emph{Proceedings of the IEEE Computer Society Conference
  on Computer Vision and Pattern Recognition}.\hskip 1em plus 0.5em minus
  0.4em\relax IEEE Computer Society, 12 2018, pp. 4510--4520.

\bibitem{j:Hernandez2021}
K.~A. Lara~Hernandez \emph{et~al.}, ``{Deep learning in spatiotemporal cardiac
  imaging: A review of methodologies and clinical usability},'' 3 2021.

\bibitem{j:Li2020}
H.~Li \emph{et~al.}, ``{Domain Generalization for Medical Imaging
  Classification with Linear-Dependency Regularization},'' \emph{arxiv}, 9
  2020.

\bibitem{j:Titano2018}
J.~J. Titano \emph{et~al.}, ``{Automated deep-neural-network surveillance of
  cranial images for acute neurologic events},'' \emph{Nature Medicine},
  vol.~24, no.~9, pp. 1337--1341, 9 2018.

\bibitem{j:Karimi2020}
D.~Karimi \emph{et~al.}, ``{Deep learning with noisy labels: Exploring
  techniques and remedies in medical image analysis},'' \emph{Medical Image
  Analysis}, vol.~65, p. 101759, 10 2020.

\bibitem{j:Zech2018}
J.~R. Zech \emph{et~al.}, ``{Variable generalization performance of a deep
  learning model to detect pneumonia in chest radiographs: A cross-sectional
  study},'' \emph{PLOS Medicine}, 2018.

\bibitem{j:Mathis2018}
A.~Mathis \emph{et~al.}, ``{DeepLabCut: markerless pose estimation of
  user-defined body parts with deep learning},'' \emph{Nature Neuroscience},
  vol.~21, no.~9, pp. 1281--1289, 9 2018.

\bibitem{j:Ronneberger2015}
O.~Ronneberger \emph{et~al.}, ``{U-Net: Convolutional Networks for Biomedical
  Image Segmentation},'' \emph{arXiv}, 5 2015.

\bibitem{j:Oktay2018}
O.~Oktay \emph{et~al.}, ``{Attention U-Net: Learning Where to Look for the
  Pancreas},'' \emph{arXiv}, 4 2018.

\bibitem{j:Branch1999}
M.~Ann~Branch \emph{et~al.}, ``{A Subspace, Interior, and Conjugate Gradient
  Method for Large-scale Bound-constrained Minimization Problems},'' \emph{SIAM
  Journal on Scientific Computing}, vol.~21, no.~1, pp. 1--23, 1999.

\bibitem{c:Leitner2021}
C.~Leitner \emph{et~al.}, ``Ustemg: an ultrasound transparent tattoo-based semg
  system for unobtrusive parallel acquisitions of muscle electro-mechanics,''
  in \emph{in Proc. 43nd Conferences of the IEEE Engineering in Medicine and
  Biology Society}.\hskip 1em plus 0.5em minus 0.4em\relax IEEE, 10 2021.

\bibitem{b:Robertson2012}
I.~H. Robertson and R.~O'Connell, ``{Vigilant attention},'' in \emph{Attention
  and Time}.\hskip 1em plus 0.5em minus 0.4em\relax Oxford University Press, 3
  2012.

\bibitem{w:tensorflow2015}
\BIBentryALTinterwordspacing
\relax Mart\'{\i}n~Abadi~et al., ``{TensorFlow}: Large-scale machine learning
  on heterogeneous systems,'' 2015, software available from tensorflow.org.
  [Online]. Available: \url{https://www.tensorflow.org/}
\BIBentrySTDinterwordspacing

\end{thebibliography}
\bibliographystyle{IEEEtran}

\end{document}